\documentclass[AMS,STIX1COL]{WileyNJD-v2}

\articletype{Article}%

\usepackage{amsmath}
\usepackage{booktabs}
\usepackage{comment}
\usepackage{bm}
\usepackage{xspace}
\usepackage{xcolor}
\usepackage{caption}
\usepackage{subcaption}

\usepackage[edges]{forest}
\usepackage{array}
\usetikzlibrary{arrows.meta}
\usetikzlibrary{shadows}
\hypersetup{hidelinks}
\usepackage[]{quoting}
\usepackage{mdframed}

\received{30 11 2020}
\revised{21 03 2021}
\accepted{Day Month Year}

\definecolor{shadecolor}{RGB}{240,240,240}
\newtheorem{exmp}{Ex.}[section]
\newenvironment{example}%
  {\begin{mdframed}[backgroundcolor=shadecolor]\begin{exmp}}%
  {\end{exmp}\end{mdframed}}

  {\begin{mdframed}[backgroundcolor=shadecolor]}%
  {\end{mdframed}}
  
\newcommand{\transp}{\mathrm{\sf T}}  

\newcommand{\eg}{\emph{e.g.},\xspace}

\newcommand{\ta}{\theta_{\text{A}}}
\newcommand{\tm}{\theta_{\text{D}}}

\newcommand{\fm}{\hat{f}_{\text{M}}}

\newcommand{\M}{M}
\newcommand{\F}{Q}
\newcommand{\FC}{Q_{\text{C}}}
\newcommand{\FI}{Q_{\text{I}}}

\newcommand{\FD}{Q_{\text{D}}}
\newcommand{\Fd}{Q_{\text{d}}}
\newcommand{\Fdi}{Q_{\text{d,I}}}
\newcommand{\FG}{Q_{\text{G}}}
\newcommand{\Fg}{Q_{\text{G}}}
\newcommand{\Fu}{Q_{\text{u}}}

\newcommand{\dv}{\delta p}

\newcommand{\C}{\mathcal{C}_i}
\newcommand{\er}{\epsilon_{\text{R}}}

\newcommand{\Nspace}{\mathsf{N}}
\newcommand{\Rspace}{\mathsf{R}}

\newcommand{\fplus}{\textcolor{teal}{\textbf{+}} \:}
\newcommand{\fminus}{\textcolor{purple}{\,\textbf{-}} \,\:}

\newcommand\fup{\textcolor{teal}{$\boldsymbol{\uparrow}$}}
\newcommand\fdown{\textcolor{purple}{$\boldsymbol{\downarrow$}}}

\newcommand\blfootnote[1]{%
  \begingroup
  \renewcommand\thefootnote{}\footnote{#1}%
  \addtocounter{footnote}{-1}%
  \endgroup
}

\begin{document}

\title{Structured learning of rigid-body dynamics:\\A survey and unified view from a robotics perspective}
\author[1]{A. René Geist}

\author[1,2]{Sebastian Trimpe}

\address[1]{\orgdiv{Intelligent Control Systems Group}, \orgname{Max Planck Institute for Intelligent Systems}, \orgaddress{\state{Stuttgart}, \country{Germany}}}
\address[2]{\orgdiv{Institute for Data Science in Mechanical Engineering}, \orgname{RWTH Aachen University}, \orgaddress{\state{Aachen}, \country{Germany}}}

\corres{A. René Geist, Heisenbergstr.\ 3, 70569 Stuttgart
Germany, \email{geist@is.mpg.de}}

%

\abstract[Abstract]{
Accurate models of mechanical system dynamics are often critical for model-based control and reinforcement learning. Fully data-driven dynamics models promise to ease the process of modeling and analysis, but require considerable amounts of data for training and often do not generalize well to unseen parts of the state space. Combining data-driven modelling with prior analytical knowledge is an attractive alternative as the inclusion of structural knowledge into a regression model improves the model's data efficiency and physical integrity. 
In this article, we survey supervised regression models that combine rigid-body mechanics with data-driven modelling techniques. 
We analyze the different latent functions (such as kinetic energy or dissipative forces) and operators (such as differential operators and projection matrices) underlying common descriptions of rigid-body mechanics. Based on this analysis, we provide a unified view on the combination of data-driven regression models, such as neural networks and Gaussian processes, with analytical model priors. Further, we review and discuss key techniques for designing structured models such as automatic differentiation.  
}


\maketitle

\blfootnote{\textbf{Abbreviations:} ML: Machine learning; GP: Gaussian Process; NN: Neural network; EOM: Equations of motion; ODE: Ordinary differential equation; \\ {\color{white}.}\hspace{2.2cm} ALM: Analytical latent modelling; ARM: Analytical (output) residual modelling; AD: Automatic differentiation; \\ {\color{white}.}\hspace{2.2cm} APN: Analytical parametric networks; COG: Center of gravity}

\section{Introduction} \label{sec:introduction}
In recent decades, increasing interest has been put on the control of mobile robots and similar agile mechanical systems. Promising frameworks for the synthesis of control policies are model-predictive control \cite{allgower2012nonlinear} and model-based reinforcement learning \cite{sutton1998introduction}. 
The performance of these algorithms rely on, or at least significantly benefit from an accurate model of the mechanical system dynamics. 
A dynamics model $\hat{f}(x,\theta)$ seeks to minimize the error to the real dynamics function $f(x)$, writing
\begin{equation} \label{eq:sim2real_gap}
    \epsilon_f(x) = f(x)-\hat{f}(x,\theta),
\end{equation}
with model parameters $\theta$ and model inputs $x$.
Many interesting mechanical systems possess complex non-linear dynamics and operate in large regions of their state-space. Therefore, numerous works on dynamics modeling choose the direct identification of the system's non-linear dynamics instead of local linear approximations. For learning a system's nonlinear dynamics, several different model types have been suggested (cf.~Fig.~\ref{fig:overview_model_types}\hspace{-0.18cm}), namely:
\begin{itemize}
    \item \emph{Analytical models (white-box models)}, which consist of functional relationships motivated by first-principles.
    \item \emph{Data-driven models (black-box models)}, which consist of solely data-driven models.
    \item \emph{Analytical structured models (gray-box models)}, which denote combinations of analytical with data-driven models.
\end{itemize}
However, the identification of the dynamics of mechanical systems from data is oftentimes challenging because real-world systems are high-dimensional and subject to complex physical phenomena such as friction and contacts. In addition, data collection on physical systems is expensive and time-consuming. These aspects in combination with the curse of dimensionality \cite[p.\,190]{nelles2013nonlinear} aggravate pure data-driven modeling. 

For these reasons, recent works aim at combining data-driven modeling techniques with a-priori available knowledge, which we will denote as \emph{structured models}.  A rich source for a-priori structural knowledge in numerous mechanical systems constitutes analytical rigid-body dynamics. 
While the bodies of a mechanical system are usually slightly elastic, it has been shown in literature that the assumption of rigid-bodies yields a useful model prior for nonlinear dynamics modeling \cite{nguyen2010using, sutanto2020encoding, hwangbo2019learning, ledezma2017first}. 


In this survey, we give a unified view on the state-of-the-art in structured learning for robotic systems using rigid-body dynamics. Structured models inherit the potential to combine the advantages of both analytical and data-driven modeling while avoiding the shortcoming of these modeling approaches when being used individually. However, a model will be rarely used in practice if its synthesis requires significant effort and its optimization is laborious. Therefore, we emphasize in this survey that recent automatic differentiation libraries such as JAX \cite{jax2018github} and PyTorch \cite{NEURIPS2019_9015} considerably simplify the synthesis of structured models and also enable GPU accelerated estimation of the model's parameters. In turn, structured modeling has the potential to emerge as one of the defining model classes that is used in the control synthesis and operation of future generations of robotic systems.
\newline\newline
Before we present our view on structured modeling of rigid-body mechanics, we first define what we understand as dynamics functions as well as discuss the strengths and shortcomings of analytical and data-driven modeling in Section \ref{sec:analytical_intro} and Section \ref{sec:data_intro} as also summarized in Table \ref{tab:pros_and_cons}\hspace{-0.18cm}.
 

\newcolumntype{C}[1]{>{\centering}p{#1}}
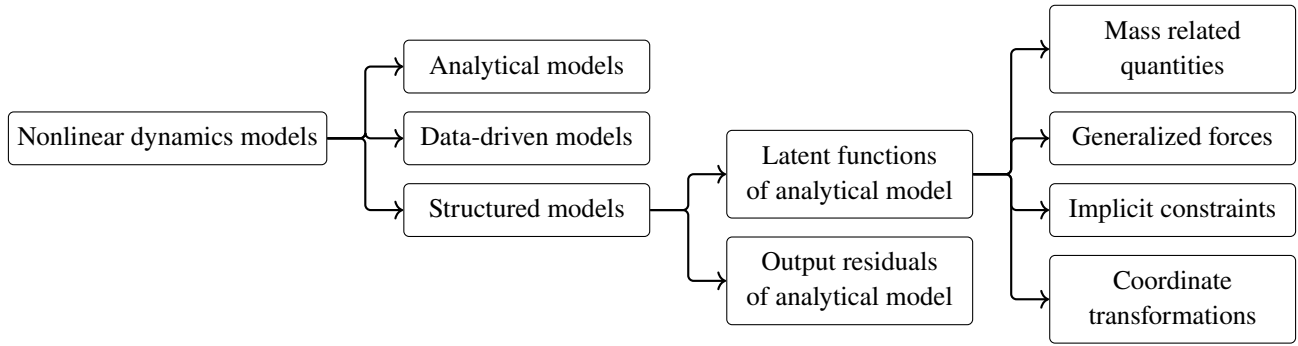
\begin{figure}[t]
\centering
\begin{forest}
  for tree={
  if level=0{align=center}{
    align={@{}C{30mm}@{}},
  },
  grow=east,
  draw,
  edge path={
    \noexpand
    \path [draw, \forestoption{edge}] (!u.parent anchor) -- +(5mm,0) |- (.child anchor)\forestoption{edge label};
  },
  parent anchor=east,
  child anchor=west,
  l sep=10mm,
  tier/.wrap pgfmath arg={tier #1}{level()},
  edge={->, thick, rounded corners=2pt},
  fill=white,
  rounded corners=2pt,
}
  [Nonlinear dynamics models
      [Structured models
      [Output residuals\\
      of analytical model]
      [Latent functions \\ of analytical model
        [Coordinate transformations]
        [Implicit constraints]
        [Generalized forces]
        [Mass related quantities]
      ]
    ]
    [Data-driven models]
    [Analytical models]
  ]
\end{forest}
\caption{Overview of different types of nonlinear dynamics models with a focus on structured modeling. The synthesis of a structured model depends on which part of an analytical model is substituted by a data-driven model. For example, a data-driven model can estimate specific forces or the entries of the inertia matrix inside of an analytical model.} \label{fig:overview_model_types}
\end{figure}

\paragraph{Nonlinear dynamics functions of mechanical systems}
The field of dynamics consists of kinematic equations describing the motion of a system via position, velocity, and acceleration, as well as kinetics, which evolves around the causal effects of generalized forces on the motion of body masses. We assume that we have a thorough understanding of the system's kinematics, which yields a generalized coordinate description as well as implicit constraint equations. When it comes to structured modeling, we are interested in utilizing knowledge on the causal relationships between the system's mass, impressed forces, and its motion. 
In most modern analytical mechanics textbooks \cite{kane1985dynamics, featherstone2014rigid}, the term \emph{dynamics} is used as a direct substitute for kinetics.

We assume that the configuration of a rigid-body mechanical system is described via the generalized position, velocity, and acceleration, which are denoted by the $n$-dimensional vectors $q(t)$, $\dot{q}(t)$, and $\ddot{q}(t)$, respectively. 
Although these vectors depend on time $t$, we generally omit the time dependence for these and other variables if this is clear from the context. Further, a specific discrete-time instant is denoted by the subscript $k$, the subsequent discrete-time step by $k+1$. In addition, the system is subject to a control input $u(t)$ or $u_k \in \mathbb{R}^{w}$. The control input induces actuation forces $\Fu(q,\dot{q},u)$ onto the system, which change the system's total energy. In the case of robots with actuated joints, $\Fu$ typically represents all forces that are caused by friction and actuation inside the joints. In this case, $\Fu$ is referred to as joint torques (usually being denoted as $\tau$). Many mechanical systems are control-affine that is $\Fu = B(q,\dot{q})u$ with $B(q,\dot{q}) \in \mathbb{R}^{n\times w}$.
For the sake of brevity, we assume that the mechanical system is \emph{fully observable}. That is, we can directly measure the state variables, or obtain a state estimate. 

The \emph{transition dynamics} $f_{\text{T}} : \mathbb{R}^{2n+w} \rightarrow \mathbb{R}^{2n}$ denotes a function from the current state-control vector to the next state, writing
\begin{align} \label{eq:transition_dynamics}
    \left[ \begin{array}{c}
         q_{k+1}\\
         \dot{q}_{k+1}
    \end{array}\right] &= f_{\text{T}}(q_{k}, \dot{q}_{k},u_k).
\end{align}
While the transition dynamics are readily amenable to data-driven models, analytical rigid-body dynamics are usually described by ordinary differential equations (ODEs).
In turn, the transition dynamics can be split up into a continuous \emph{forward dynamics} function  $f_{\ddot{q}}: \mathbb{R}^{2n+w} \rightarrow \mathbb{R}^{n}$, which maps the state-control vector to the system's acceleration, 
\begin{align}
\ddot{q} &= f_{\ddot{q}}(q, \dot{q},\Fu),
\end{align}
and a numerical integration method predicting the next state using $\ddot{q}$. 
Commonly used integration methods such as Runge-Kutta-45 predict the next state as a function of several state-acceleration vectors at different time steps \cite{davis2007methods}.

Alternatively, one can rearrange the forward dynamics to obtain the system's \emph{inverse dynamics} function $f_{u}: \mathbb{R}^{3n} \rightarrow \mathbb{R}^{n}$, which computes the (feed-forward) generalized actuation force that is required to achieve a state-acceleration vector, writing
\begin{align} \label{eq:inverse_dynamics}
\Fu &= f_u(q, \dot{q}, \ddot{q}).
\end{align}
 The inverse dynamics mapping in \eqref{eq:inverse_dynamics} can be non-injective. In this case, the identification of inverse dynamics via supervised learning models requires additional mathematical techniques \cite{jordan1992constrained}, which are not discussed in this survey. 
\newline\newline
To simplify the discussion, we use the term \emph{dynamics} interchangeably for functions modeling either the transition, forward, or inverse dynamics functions of a mechanical system. To this end, we consider the dynamics of a mechanical system to be described by the general function $f: \mathbb{R}^{n_x} \rightarrow \mathbb{R}^{n_y}$ where $f(x)=y$ with input vector $x \in \mathbb{R}^{n_x}$ and output vector $y \in \mathbb{R}^{n_y}$.  For example, if $f$ denotes the system's forward dynamics, then $x$ is a state-control vector and $y$ the system's acceleration. 

\paragraph{Actuation dynamics}
In the control of a mechanical system, one usually calculates the desired actuation force $\F_{\text{u,desired}}$ that is sent to computational routines which in return cause the actuators to impress $\Fu$ onto the system.
Yet, $\F_u$ likely deviates from $\F_{\text{u,desired}}$ due to friction and other unmodelled physical phenomena inside the actuators, flexibilities in gear-belts or attached shafts, free-play in an attached transmission, and internal dynamics of cascaded feedback control loops.
To account for this difference, one can define the \emph{actuation dynamics} function $f_{\text{ACT}} : \mathbb{R}^{n_\rho} \rightarrow \mathbb{R}^{n}$ as
 \begin{equation}
     \Fu' = \Fu(\rho, \F_{\text{u,desired}}) - \F_{\text{u,desired}}(\rho) = f_{\text{ACT}}(\rho, \F_{\text{u,desired}}),
 \end{equation}
 where $\rho \in \mathbb{R}^{n_\rho}$ denotes a suitably chosen vector of observables such as state data $\{q,\dot{q}\}$ from several time steps, the error to a target state, or the current applied to the actuators.
 For example, the generation of torque inside a robot arm's brush-less motors requires 
 that a $\F_{\text{u,desired}}$ is transformed by low-level control routines into motor currents that cause a motor torque; the motor torque is altered by a potential gear train as well as joint friction; and finally $\Fu$ is measurable in the joints.
 While the actuation dynamics can play a significant role in practical robot implementations, rigid-body dynamics typically starts at the level of forces. In this survey, we mostly focus on structured learning of the rigid-body dynamics. 

\paragraph{Supervised regression}
In this survey, we assume that a model $\hat{f}(x;\theta)=\hat{y}$ with model parameters $\theta$ shall be trained on informative data  
\begin{equation}
   \mathcal{D} \,\widehat{=}\, \{\tilde{x}_k,\tilde{y}_k\}_{k=1}^N 
\end{equation}
 where $N$ denotes the number of data points, $\tilde{x}$ denotes noisy measurements of $x$, and $\tilde{y}$ denotes noisy measurements of $f$.  The term training or learning refers to the estimation of $\theta$ from data. The term supervised refers to a model being trained on input-output data pairs of the dynamics function.

\subsection{Analytical models} \label{sec:analytical_intro}
Analytical models are commonly used in the identification of the dynamics of mechanical systems \citep{sutanto2020encoding, atkeson1986estimation, ting2006bayesian, traversaro2016identification, wensing2017linear, ledezma2018fop}. 
Analytical models of rigid-body dynamics are based on physical axioms and principles underlying the motion of mechanical systems, \eg Newton's axioms of motion. These axioms and principles spawn dynamics equations such as the Newton-Euler equations and the Euler-Lagrange equations. The equations of analytical mechanics are combinations of numerous physically motivated functions such as coordinate-transformations, forces, and the inertia matrix which depend on physical parameters. Physical parameters are for example the mass of a rigid-body, a friction coefficient, the length of a kinematic link, or the stiffness coefficient of a spring. The numerous functions that form an analytical model can usually not be observed individually which is why we refer to them as being \emph{latent}. Notably, the latent functions and parameter estimates of analytical models yield a physical interpretation that can guarantee \emph{out of sample generalization} (i.e., the validity of the model in regions where no data has been observed) and earns them the name white-box models. However, on real systems, physical phenomenons such as \emph{friction}, \emph{damping}, and \emph{contacts} aggravate an accurate identification of an analytical dynamic model. In addition, other physical phenomena such as \emph{body elasticities} cannot be modeled by analytical rigid-body dynamics in the first place and therefore lead to errors. 
In practice, some of the parameters of an analytical model are estimated from data using linear regression or gradient-based optimization. 
However, the analytical model errors and the limited representative power of analytical force models can lead to physically inconsistent parameter estimates such as negative body masses \cite{ting2006bayesian}. Analytical models whose parameters are estimated via gradient-based optimization could be seen as parametric data-driven models which is why we refer to them as analytical parametric networks (APN). Analytical parametric networks provide profound insights for the synthesis of analytical structured models, which is why we briefly discuss these works in Section \ref{sec:APN}.

\begin{table}
        \caption{Pros and cons of analytical rigid-body models compared to data-driven models.} \label{tab:pros_and_cons}
        \centering
        \begin{tabular}{@{}ll@{}}
        \toprule
        \textbf{Analytical models} & \textbf{Data-driven models} \\ \midrule
        \fplus Data-efficient & \fminus Data-hungry \\
        \fplus Human-interpretable & \fminus Usually black-box model \\
        \fplus Out-of-sample generalization & \fminus Usually data-point interpolation \\
        \fminus Large prediction errors & \fplus Less restrictive assumptions \\
        \fminus Full prior knowledge required  & \fplus Less prior knowledge required \\ 
        \,$\circ$\: Deterministic & \,$\circ$\: Possibly probabilistic \\
        \bottomrule
        \end{tabular}
\end{table}

\subsection{Data-driven models} \label{sec:data_intro}
When analytical modeling is not feasible (e.g, because the physical process is unknown), or the effort required to obtain a sufficiently small prediction error appears to be too large, the nonlinear dynamics can be directly learned from data. The design of the data-driven model places prior assumptions on the functions it can approximate. 
To reduce the prediction error of these models, one needs to estimate their parameters, which is referred to as training. Parameters can be either the hyperparameters of the kernel of a non-parametric method, such as a Gaussian process (GP), or the parameters of a parametric method such as a neural network (NN). As the same optimization methods are commonly used for training parametric models as well as GPs, we refer in what follows to hyper-parameters also as parameters. However, training hyperparameters of a kernel determines an entire population of features while training the parameters of a parametric model usually determines a function approximation.

Data-driven models contrast analytical models as most often their parameter estimates do not yield a physically insightful interpretation.
Moreover, the sheer amount of parameters and interlinked latent functions in most data-driven models such as neural networks eludes human comprehension. Thus they are commonly referred to as black-box models.

One of the largest concern with the identification of dynamics via data-driven models is \emph{sample complexity}; that is, the number of training points a data-driven model requires to successfully approximate a function. On physical systems, data is often sparse because the data collection is subject to life-time and cost constraints. While it has been proposed to identify structural knowledge directly from data \cite{SahooLampertMartius2018:EQLDiv, baumann2020identifying}, the identification of structure requires significant amounts of data. In the next section, we detail our understanding of structure, and why it is beneficial to combine analytical structure with data-driven modeling.   

\subsection{Structured models}
In this survey, we interpret the term structure as either mathematical properties of functions, or alternatively, causal dependencies between functions. In the field of nonlinear system identification, a \emph{gray-box model} denotes the combination of analytical with data-driven modeling \cite{nelles2013nonlinear, lennart1999system}. In slight contrast, a structured model denotes the combination of data-driven models with \emph{some form of} structural prior knowledge that reduces the required complexity of the model class (cf. \cite[p.\,192]{nelles2013nonlinear}). Such structural knowledge can be an analytical model or alternatively, the way in which different data-driven models are being connected with each other. Therefore, a gray-box model is a structured model, but not all structured models are gray-box models. As suggested by \citet{nelles2013nonlinear}, structured models can be further categorized as:
\begin{itemize}
    \item \emph{Hybrid structures}: The combination of different sub-models to a single structured model.\footnote{The term hybrid structures/models should not be confused with hybrid systems which commonly denote the combination of continuous and discrete-time systems.} This model type is further divided into:
    \begin{itemize}
    \item \emph{Parallel model}: The outputs of all sub-models are summed up to yield the output of the hybrid model.
    \item \emph{Series model}: The sub-models are connected in series, that is the output of one sub-model forms the input to another sub-model and so forth.
    \item \emph{Parameter scheduling model}: The parameters of a sub-model are scheduled by the output of another sub-model.
\end{itemize}
\item \emph{Projection-based structures}: The input space is projected into a lower-dimensional latent space.
\item \emph{Additive structures}: The input dimensions are grouped into several sub-spaces which act as input to sub-models. The additive model's output is the sum of the sub-models' outputs.
\item \emph{Hierarchical structures}: A model is composed of parallel, series, and additive structures yielding hierarchical causal relationships between the inputs and outputs of the sub-models.
\item \emph{Input space decompositions}: The input space is decomposed into several regions that each constitutes a sub-space to a sub-model. Note that the input space is split up region-wise rather than dimension-wise as in additive structures. 
\end{itemize}
Notably, analytical rigid body dynamics can yield many of the above types of structural knowledge, which makes it a versatile toolbox for structured learning.

In the field of robotics, the term structured learning usually refers to the combination of analytical mechanics with data-driven modeling \cite{lutter2018deep, gupta2020structured, geist2020gp2}. 
The term structured mechanics model has also been used in robotics literature \cite{gupta2020structured}.
As it is the analytical prior knowledge that structures a data-driven model, we use the term \emph{analytical structured model} to refer to a structured model that uses analytical rigid-body mechanics as a model prior to data-driven modeling.

\paragraph{A unified view on analytical structured models}
The core idea of analytical structured modeling is to complement an analytical model via data-driven modeling to infer its errors from data. 
An analytical model $\hat{f}_{\text{A}}(x;\ta)$ with parameters $\ta$ defines a hierarchical structure between several latent functions such as forces and inertias. In principle, one can directly use the analytical model to faithfully estimate $\ta$ by minimizing a suitable metric $\|\tilde{y}-\hat{f}_{\text{A}}(\tilde{x};\ta)\|$. 
However, an analytical model is only an approximation of the dynamics $f(x)$ such that
\begin{equation} \label{eq:analytical_approximation_error}
    f(x) = \hat{f}_{\text{A}}(x;\ta) + \epsilon_{\text{A}}(x;\ta),
\end{equation}
with the anayltical model errors $\epsilon_{\text{A}}(x)$.
To reduce $\epsilon_{\text{A}}$ and potentially improve the estimate of $\ta$, one can add a data-driven model $\hat{\epsilon}_{\text{A}}(x;\tm)$ with parameters $\tm$ to the analytical model, writing
\begin{equation} \label{eq:residual_model}
        \hat{f}(x;\ta,\tm) = \hat{f}_{\text{A}}(x;\ta) + \hat{\epsilon}_{\text{A}}(x;\tm).
\end{equation}
We refer to the parallel model structure in \eqref{eq:residual_model} as \emph{analytical (output) residual modeling} (ARM). Usually, the measurements obtained from the real dynamics are subject to additional measurement noise and bias $\epsilon_y(\tilde{x})$, such that a system measurement follows from $\tilde{y}=\hat{f}_{\text{A}}(\tilde{x};\ta) + \epsilon_{\text{A}}(\tilde{x})+\epsilon_y(\tilde{x})$. Therefore, the data driven model usually models the residual $\epsilon_{\text{A}}(\tilde{x})+\epsilon_y(\tilde{x})$ jointly. ARM forms a natural extension to analytical modeling and is conceptually easy to implement (while there can be of course practical challenges such as noise). 
However, if $\hat{f}_{\text{A}}(x;\ta)$ is inaccurate, the data-efficiency of the resulting structured model suffers significantly. In what follows, we omit the arguments in the notation of a function if these are clear from context.

Consequently, the question arises on how we can improve the prediction accuracy of $\hat{f}_{\text{A}}$ itself by using a data-driven model. 
An important insight for structured learning is that the cause of parts of the analytical model error $\epsilon_{\text{A}}$ lies hidden in the functions that form the analytical model  $\hat{f}_{\text{A}}$. 
For example, parts of $\epsilon_{\text{A}}$ can be caused by an inaccurate kinematics model or an inaccurate model for the friction forces. Therefore, instead of estimating directly $\epsilon_{\text{A}}$, one can also substitute unknown latent functions or their residuals with a data-driven model \emph{inside}  $\hat{f}_{\text{A}}$  (to be made precise in Section \ref{sec:structured_learning}).
We refer to a structured model in which data-driven models substitute/augment parts of $\hat{f}_{\text{A}}$ as \emph{analytical latent modeling} (ALM).

\subsection{Overview and notation}
The remainder of this article is organized as follows. 
Section \ref{sec:mechanics} introduces the reader to the analytical mechanics of rigid-body systems. This section introduces different types of structural knowledge while emphasizing the importance of linear operators in rigid-body mechanics.
While we do not aim at a 
comprehensive overview of rigid-body dynamics herein,
Section \ref{sec:mechanics} 
outlines important aspects of analytical mechanics that are useful for the synthesis of structured models and needed for the purpose of this survey.
Section \ref{sec:structured_learning} discusses the current state of the art in analytical structured models. Via the discussion of the previous sections, we develop a unified view on ARM as well as ALM.
Section \ref{sec:synthesis} details the key mathematical techniques required to combine the most common data-driven models, namely GPs and NNs, with analytical equations. Further, we emphasize the importance of recent developments on automatic differentiation libraries for a straightforward design and training of analytical structured models. 
Finally, we discuss a case study of modeling the dynamics of a robot arm to illustrate the presented concepts.
\newline
\newline
In this work, we adapt the following notation. A unit matrix is denoted by $I$, and a matrix of zeros by $\boldsymbol{0}$. The vertical concatenation of vectors $\{v_1, \dots, v_n\}$ is denoted as $\text{vec}\{v_1, \dots, v_n\}\,\widehat{=}\,[v_1^\transp \dots v_n^\transp]^T$. The null space of a matrix $A: \mathbb{R}^{m \times n}$ is defined as $\Nspace(A) = \{x \in \mathbb{R}^{n}: Ax=0\}$, its range space as $\Rspace(A) = \{y \in \mathbb{R}^{m}: \exists x \in \mathbb{R}^{n} \text{ such that }  y = Ax\}$, and further we have that $\mathbb{R}^{n} = \Rspace(A^T) \oplus \Nspace(A)$ and $\mathbb{R}^{m} = \Rspace(A) \oplus \Nspace(A^T)$ \citep{beard2002linear}. If a matrix $A(x): \mathbb{R}^{m\times n}$ with $m \leq n$ and $\text{rank}(A)=m$ then $\Nspace(A^\transp)=\{0\}$. $A^+$ denotes the Moore-Penrose pseudo (MP) inverse of $A$. 
A vector $x \in \mathbb{R}^{n}$ can be split up in a range space part $x_{\Rspace(A^\transp)}$ and a null space part $x_{\Nspace(A)}$, such that $x=x_{\Rspace(A^\transp)}+x_{\Nspace(A)}$. The projections of a vector into $\Nspace(A)$ and $\Rspace(A^\transp)$, (cf. \cite{beard2002linear}), are given by
\begin{equation}
    P^{\Nspace(A)} = I-A^+A, \:\text{ and }\: P^{\Rspace(A^\transp)} = A^+A,
\end{equation}
such that $x_{\Rspace(A^\transp)}=P^{\Rspace(A^\transp)}x$ and $x_{\Nspace(A)}=P^{\Nspace(A)}x$.
Generalized forces are denoted by $Q_{(\cdot)}$, dimensions by $n_{(\cdot)}$, error functions by $\epsilon_{(\cdot)}$, dynamics functions as $f_{(\cdot)}$, dynamics models as $\hat{f}_{(\cdot)}$, and model parameters by $\theta_{(\cdot)}$ where the respective indice ``$(\cdot)$'' specifies a particular type. 


\section{A glimpse on analytical rigid-body mechanics} \label{sec:mechanics}
%

%
In this section, we give an overview on important aspects of rigid-body dynamics which provides the means to discuss current literature on analytical structured modeling. Here, 
we show \emph{one} possible way on how the system's equations of motion (EOM) can emerge from the interplay of kinematics, dynamic principles, and constraint equations. 

We first limit the discussion to the Newton-Euler equations of a system of $N_{\text{b}}$ rigid bodies subject to holonomic constraints. By using a set of independent generalized coordinates we eliminate the constraint forces from the EOM. Then, we briefly discuss how in the Euler-Lagrange equations certain forces can be modelled in terms of potential functions. These dynamics formulations are frequently used for the description of robot dynamics \cite{siciliano2010robotics} and the simulation of rigid-body systems \cite{featherstone2014rigid}. Afterwards, we discuss how additional implicit constraints are incorporated into the EOM. With the discussion of explicit and implicit constraints, we illustrate how constraint equations provide structural knowledge on the vector spaces in which certain forces are bound to lie.


\paragraph*{Kinematics}
To derive the EOMs of a multi-body system, the motion variables -- position, velocity and acceleration -- of its $N_b$ rigid bodies must be described with respect to an inertial frame. 
The position of every point of the $i$-th rigid body can be described by a position vector $r_i(t)\in\mathbb{R}^3$ pointing to the origin of a body-fixed frame with respect to an inertial frame and a rotation matrix $R_i(t) \in SO(3)$ describing the rotation of the body-fixed coordinate frame with respect to the inertial frame. $SO(3)$ denotes the subgroup of orthogonal matrices of size three with determinant $+1$. 
Subsequently, the multi-body system has at most $6N_\text{b}$ degrees of freedom (DOF). The velocity of the body is described by the the translational velocity $\dot{r}_i(t)$ and the rotational velocity $\omega_i(t)$. The rotational velocity $\omega_i(t)=\dot{\varphi}_i=[\omega_1, \omega_2, \omega_3]^\transp$ is obtained in terms of $R(t)$ (cf. \cite[p.\,28]{schiehlen2014applied}) as
\begin{equation}
    \text{crossp}\{\omega\} = \dot{R}(t)R(t)^\transp,
\text{ with } \text{crossp}\{\omega\}=\begin{bmatrix}0 & -\omega_3 & \omega_2 \\ \omega_3 & 0 & -\omega_1 \\ -\omega_2 & \omega_1 & 0 \end{bmatrix},
\end{equation}
with the infinitesimal instantaneous rotation vector $\varphi_i \in \mathbb{R}^3$ \cite[p.\,67]{woernlemehrkorpersysteme}.  

Instead of expressing $r_i(t)$ in Cartesian coordinates, it is often more practical to formulate $r_i(t)$ as well as $R_i(t)$ in terms of a generalized coordinate vector $q(t)$.
For example, one could use spherical coordinates to denote a point in space (cf. \cite[p.\,14]{schiehlen2014applied}) or Cardano angles to describe rotations (cf. \cite[p.\,24]{schiehlen2014applied}). The vector $q$ is termed \emph{minimal} if it consists of independent coordinates that equal the system's DOF.

\subsection{Newton-Euler equations: Structural \label{sec:mechanics_newtonian} knowledge between forces and acceleration}
To keep the discussion concise, we make the following assumptions for Section \ref{sec:mechanics_newtonian} and Section \ref{sec:lagrange_equations}:
\begin{itemize}
    \item The EOM of a system of $N_\text{b}$ rigid bodies are expressed in terms of minimal generalized coordinates.
    \item The effect of the constraint forces onto the system are modelled through $n_\text{E}$ independent holonomic constraint equations.    
    \item  The origin of the body-fixed frame lies at the body's centre of gravity (COG).
\end{itemize}
While these assumptions are common for the derivation of robot dynamics, other assumptions such as using an accelerated frame of reference or a reference system with its origin being not placed in the COG are also used. A more far-reaching description of multibody dynamics is given in \cite{schiehlen2014applied}.

 The Newton-Euler EOM of the i-th rigid body with respect to the COG read
\begin{equation}
   \tilde{M}_i
   \ddot{p}_i = 
     F_{\text{C},i} + F_{\text{e},i} + F_{\text{E},i} \label{eq:newton_roots}
\end{equation}
with $\ddot{p}_i = \text{vec}\{\ddot{r}_i, \dot{\omega}_i\}$, the block-diagonal matrix as $\tilde{M}_i=\textbf{diag}\{m_i I_3, \Theta_{S,i}\}$ consisting of the body's mass $m_i$ and its inertia matrix $\Theta_{S,i}$, the bias vector $F_{\text{C},i} = \text{vec}\{\boldsymbol{0}, -\tilde{\omega}\Theta_{S,i}\omega_i \}$, the impressed forces and torques $F_{\text{e},i} = \text{vec}\{f_{\text{e},i}, \tau_{\text{e},i}\}$, and the explicit constraint forces and torques $F_{\text{E},i} = \text{vec}\{f_{\text{E},i}, \tau_{\text{E},i}\}$ \citep[p.\,76]{schiehlen2014applied}. 

\paragraph{Principle of virtual work and D'Alembert-Lagrange's principle}
To eliminate the constraint forces and torques from \eqref{eq:newton_roots} it is assumed that the constraint forces are \emph{ideal}, that is, the constraint forces do zero work under virtual displacements $\delta r_i \in \mathbb{R}^3$ and the reaction torques do zero work under virtual rotations $\delta \varphi_i \in \mathbb{R}^3$. In turn, with $\dv_i = \text{vec}\{\delta r_i, \delta \varphi_i\}$, it is postulated that the ideal constraint forces $F_{\text{E},i}$ respect the following inner product
   \begin{equation} \label{eq:principle_of_virtual_work}
       \sum_{i=1}^{N_{\text{b}}} \dv_i^\transp F_{\text{E},i} = \dv^\transp F_{\text{E}} = 0, 
    \end{equation}
with the $\dv_i$ of all bodies being denoted jointly as $\dv = \text{vec} \{\dv_1, \dv_2, \dots \dv_{N_\text{b}}\}$ as well as the explicit constraint forces of all bodies being denoted as $F_{\text{E}} = \text{vec} \{F_{\text{E},1}, F_{\text{E},2}, \dots , F_{\text{E},N_{\text{b}}}\}$. Note that, if the $i$-th body is not subject to any constraint forces then $F_{\text{E},i}=0$.
The vectors of virtual displacements and virtual rotations denote infinitesimal vectors that are compatible with the constraints while by definition not varying the time variable (cf. \citep[p.\,133]{udwadia2007analytical}, \citep[p.\,85]{schiehlen2014applied}), see also Section \ref{sec:constraint_mechanics} for a more in depth discussion. 
The D'Alembert-Lagrange principle \citep{d1743traite, de1788mechanique} in its to multibody systems extended form \citep[p.\,92]{schiehlen2014applied} is obtained by multiplication of \eqref{eq:newton_roots} from the left by $\dv_i^\transp$ and applying \eqref{eq:principle_of_virtual_work}, such that
\begin{equation}
   \sum_{i=1}^{N_{\text{b}}}  \dv_i^\transp \left(\tilde{M}_i  \ddot{p}_i
   - F_{c,i} -
   F_{e,i} \right) = 0. \label{eq:dalembert1}
\end{equation}
Due to the presence of $F_\text{E}$, the components of $\dv_i$ are dependent on each other. 
In what follows, we outline how to obtain the EOM from \eqref{eq:dalembert1} by resorting to explicit constraint equations that are expressed in terms of a minimal set of generalized coordinates.

\newpage
\paragraph{Explicit holonomic constraints and generalized coordinates}
In multi-body systems, the rigid bodies are usually subject to kinematic mechanisms which apply constraint forces onto the multibody system. The constraint forces reduce the system's DOF. The effect of the constraint forces on the system's kinematics can be modelled via constraint equations. In this work, the term constraints refers to algebraic equations that describe the system's admissible states \cite[p.\,44]{featherstone2014rigid}. Constraints can be either holonomic or nonholonomic. A constrained system is holonomic if its position variables are integrals of the velocity variables; otherwise, the system is nonholonomic \cite[p.\,41]{featherstone2014rigid}. We briefly discuss nonholonomic systems in Section \ref{sec:constraint_mechanics}. 

As the system has $n_\text{q} = 6N_\text{b}-n_\text{E}$ DOF and $q \in \mathbb{R}^{n_\text{q}}$, the $i$-th bodies position and orientation can be written in terms of explicit constraints as
\begin{align} \label{eq:generalized_explicit}
    r_i(t) &= r_i(q,t),\hspace{1cm}  R_i(t) = R_i(q,t).
\end{align}
Note that a free system (that is no constraint forces act onto the bodies) can be seen as a special case of a holonomic constrained system in which $n_\text{E}=0$, $q$ is of dimension $6N_\text{b}$, and \eqref{eq:generalized_explicit} denotes a suitable coordinate transformation \cite[51]{schiehlen2014applied}. Differentiation of \eqref{eq:generalized_explicit} with respect to time yields constraint equations for the systems velocities and accelerations as
\begin{alignat}{2}
    \dot{r}_i(q,\dot{q},t) &= J_{r,i} \dot{q} + \frac{\partial r_i}{\partial t}, \hspace{3cm} 
    \omega_i(q,\dot{q},t) &&= J_{R,i}\dot{q} + \frac{\partial \varphi_i}{\partial t} ,\label{eq:explicit_vel}\\
    \ddot{r}_i(q,\dot{q}, \ddot{q})  &= J_{r,i} \ddot{q} + \dot{J}_{r,i} \dot{q} + \frac{\partial \dot{r}_i}{\partial t}, \hspace{2cm}
    \dot{\omega}_i(q, \dot{q}, \ddot{q}) &&= J_{R,i}\ddot{q} + \dot{J}_{R,i} \dot{q} + \frac{\partial \dot{\varphi}_i}{\partial t}, \label{eq:explicit_acc}
\end{alignat}
with the translational Jacobian matrix $J_{r,i}(q,t) \in \mathbb{R}^{3 \times n_\text{q}}$ and the rotational Jacobian matrix $J_{R,i}(q,t) \in \mathbb{R}^{3 \times n_\text{q}}$ \cite[p.\,195]{woernlemehrkorpersysteme}. Note while $J_{r,i}(q,t)= \frac{\partial r_i(q)}{\partial q}$, the matrix $J_{R,i}(q,t) = \frac{\partial \varphi_i(q)}{\partial q}$ is often obtained by directly expressing $\omega_i$ in terms of $\dot{q}$ \cite[p.\,33]{schiehlen2014applied}. One can rewrite \eqref{eq:explicit_acc} more compactly as
\begin{equation}
    \ddot{p}_i(q, \dot{q}, \ddot{q}) = J_{i} \ddot{q} + \tilde{J}_i, 
\end{equation}
with the Jacobian matrix $J_i = [J_{r,i}^\transp, J_{R,i}^\transp]^\transp$ and $ \tilde{J}_i = \text{vec}\{\dot{J}_{r,i} \dot{q} + \frac{\partial \dot{r}_i}{\partial t}, \,\dot{J}_{R,i} \dot{q} + \frac{\partial \dot{\varphi}_i}{\partial t}\}$.
Importantly, the virtual vector $\dv_i$ can be expressed in terms of the generalized virtual displacement vector $\delta q \in \mathbb{R}^{n_q}$ using \eqref{eq:explicit_vel} (cf. \citep[p.\,50]{schiehlen2014applied}) such that
\begin{equation} \label{eq:virtual_transfo}
   \dv_i = J_i \delta q.
\end{equation}
By inserting  \eqref{eq:virtual_transfo} into \eqref{eq:principle_of_virtual_work} one obtains
\begin{equation} \label{eq:JTFiszero}
           \sum_{i=1}^{N_{\text{b}}} \delta q^\transp J^\transp_i F_{\text{E},i} = 
           \delta q^\transp J^\transp F_{\text{E}} = 0,
\end{equation}
with $J = [J_1^\transp, J_2^\transp, \dots J_{N_\text{b}}^\transp]^\transp$. As \eqref{eq:JTFiszero} must hold for an \emph{arbitrary} $\delta q$, we also have $J^\transp F_{\text{E}} = 0$ such that
\begin{equation} \label{eq:space_of_Fe}
    F_{\text{E}} \in \Nspace(J^\transp).
\end{equation}
Further, with \eqref{eq:space_of_Fe} and $\mathbb{R}^{6N_\text{b}} = \Rspace(J) \oplus \Nspace(J^\transp)$ (cf. \cite{beard2002linear}), from \eqref{eq:principle_of_virtual_work} follows
\begin{equation} \label{eq:space_of_dv}
    \dv \in \Rspace(J).
\end{equation}

%
\paragraph{Equations of motion with explicit constraints}
By transformation of \eqref{eq:dalembert1} into generalized coordinate form using \eqref{eq:generalized_explicit}, \eqref{eq:explicit_vel}, \eqref{eq:explicit_acc}, and \eqref{eq:virtual_transfo}, one obtains
\begin{equation} \label{eq:generalized_dalembert_step}
   \delta q^\transp  \sum_{i=1}^{N_{\text{b}}}  J^\transp_i \left(\tilde{M}_i  (J_i \ddot{q} + \tilde{J}_i)  
   - F_{\text{C},i} - F_{\text{e},i} \right) = 0.
\end{equation}
As \eqref{eq:generalized_dalembert_step} must hold for an arbitrary $\delta q$, the local EOM of the $i$-th rigid body are obtained as
    $M_i \ddot{q} =  Q_{C,i} + Q_{e,i},$
with $M_i(q,t)=J_i^\transp \tilde{M}_i J_i$,  $Q_{\text{C},i}(q,\dot{q},t) = J_i^\transp (F_{\text{C},i}(q,\dot{q},t) - \tilde{M}_i \tilde{J}_i)$, and $Q_{\text{e},i}(q,\dot{q},t)=J_i^\transp  F_{\text{e},i}$  \cite[p.\,100]{schiehlen2014applied}. 
In return, one obtains the EOM of the  multibody system as
    \begin{equation} \label{eq:newton_euler}
                 \M\ddot{q} = \F,
    \end{equation}
with  $\F = \FC + \F_{\text{e}}$, the generalized inertia matrix  $M=\sum_{i=1}^{N_\text{b}}M_i$, the generalized bias force  $\FC = \sum_{i=1}^{N_\text{b}} \F_{\text{C},i}$, and the generalized impressed force $\F_{\text{e}} =\sum_{i=1}^{N_\text{b}} \F_{\text{e},i}$ \cite[p.\,107]{schiehlen2014applied}. As pointed out by \citet[p,\,40]{featherstone2014rigid}, if the impressed forces $\F_{\text{e}} = - \FC$ then the system's acceleration amounts to null.
We assume that the generalized coordinates are chosen such that $M(q,t)$ and its inverse $M^{-1}(q,t)$ are symmetric and positive-definite, writing $\ddot{q}^{\top}M(q,t)\ddot{q} > 0$. Then, $M^{-1}(q,t)$ maps the $n_q$-dimensional generalised force space to the $n_q$-dimensional generalized acceleration space.  
As the impressed forces possess different properties depending on their source of origin, we further split up the force vectors as
    \begin{align} \label{eq:newton_euler_exapnded}
\F_{\text{e}}= \F_G + \F_D \: \text{ with }\: \FD = \Fd + \Fu.
    \end{align}
The conservative force $\FG(q, \dot{q})$ arises from a physical potential $V(q,\dot {q})$, such as a spring or a gravitational force. The non-conservative force vector $\FD(q, \dot{q},t)$ denotes forces that can change the system's total energy via external forces $\Fd(q, \dot{q},t)$ and the actuation forces $\Fu(q, \dot{q},t)$, respectively. Note that $\Fd$ is often a dissipative force, that is, it can only reduce the system's total energy. If the system is fully actuated, one obtains the system's inverse dynamics by rearranging \eqref{eq:newton_euler} as
\begin{equation}
    \Fu = M \ddot{q} - \FC - \FG - \Fd. \label{eq:invere_newton}
\end{equation}
 Note that there exist computationally efficient recursive algorithms for the derivation of the EOM of many types of robotic systems. An introduction to recursive rigid-body algorithms is given in \cite{featherstone2014rigid}.

\subsection{Euler-Lagrange equations: Structural knowledge for energy conservation} \label{sec:lagrange_equations}
In this section, we detail how some of the terms in \eqref{eq:newton_euler_exapnded} can be obtained in terms of potential functions.
The resulting equations can be used to include energy conservation into an analytical structured model as discussed in Section \ref{sec:latent_energy_conservation}. A detailed introduction to the Lagrange equations is given in \cite[p.\,68]{layton2012principles} and more specifically for robot arms in \cite[p.\,247]{siciliano2010robotics}.  

The Lagrangian function $\mathcal{L}(q, \dot{q})$ can be obtained as the difference between the kinetic energy $T(q, \dot{q})$ and the potential energy $V(q, \dot{q})$, writing
\begin{equation} \label{eq:general_lagrangian}
    \mathcal{L} = T - V = \frac{1}{2}\dot{q}^{\top} M \dot{q} - V.
\end{equation}
In return, the Euler-Lagrange equation of a rigid-body system's $i$-th dimension can be derived via the calculus of variations as
\begin{equation} \label{eq:lagrangian_oned}
\frac{d}{dt} \frac{\partial \mathcal{L}}{\partial \dot{q}_i} - \frac{\partial \mathcal{L}}{\partial q_i}= 0.
\end{equation}
Note that the Lagrangian differs from the system's total energy $E = T + V$. The Lagrangian is used to mathematically express that a conservative system must take a path of stationary action via \eqref{eq:lagrangian_oned} such that $E$ remains constant. 
Rewriting \eqref{eq:lagrangian_oned} in vector notation and adding the non-conservative forces $\FD$ yields
\begin{equation} \label{eq:euler_lagrange}
\frac{d}{dt} \nabla_{\dot{q}}\mathcal{L} - \nabla_{q} \mathcal{L}=\FD,
\end{equation}
with $(\nabla_{\dot{q}})_i = \frac{\partial}{\partial \dot{q}_i}$.
One can apply the chain rule to expand the \emph{time-derivative} of the Lagrangian's partial derivative as 
\begin{equation} \label{eq:time_parametrization}
    \frac{d}{dt} \nabla_{\dot{q}}\mathcal{L} = \left(\nabla_{\dot{q}}\nabla_{\dot{q}}^{\top} \mathcal{L}\right) \ddot{q} + \left(\nabla_{q}\nabla_{\dot{q}}^{\top} \mathcal{L}\right) \dot{q},
\end{equation} 
with the $n \times n$ matrix $\left(\nabla_{q}\nabla_{\dot{q}}^{\top} \mathcal{L}\right)_{ij} = \frac{\partial^2 \mathcal{L}}{\partial q_i\partial \dot{q}_j}$. Therefore, one obtains \eqref{eq:newton_euler} in terms of the Lagrangian as
\begin{equation} \label{eq:fd_cranmer}
    \ddot{q} = \left(\nabla_{\dot{q}}\nabla_{\dot{q}}^{\top} \mathcal{L}\right)^{-1} \left( - \left(\nabla_{q}\nabla_{\dot{q}}^{\top} \mathcal{L}\right) \dot{q} + \nabla_{q} \mathcal{L} + \FD\right).
\end{equation}
In what follows, it is assumed that $V(q)$ only depends on $q$ to keep the expressions concise. The previous equation can also be written in terms of $M = \nabla_{\dot{q}}\nabla_{\dot{q}}^{\top} \mathcal{L}$ and $\FG = - \nabla_{q} V(q)$ to yield an alternative description of the forward dynamics as
\begin{equation} \label{eq:fd_lutter}
\ddot{q}=M^{-1}\left( - \nabla_{q} (\dot{q}^{\top}M)\dot{q}+  \frac{1}{2}\nabla_{q} \left(\dot{q}^{T} M \dot{q}\right) - \nabla_{q} V + \FD \right).
\end{equation}
The above equations yield parametrizations of the fictitious force as
\begin{align} \label{eq:fictitious_force}
   \FC &= - \left(\nabla_{q}\nabla_{\dot{q}}^{\top} T\right) \dot{q} + \nabla_{q}T
   = - \nabla_{q}(\dot{q}^{\top}M) \dot{q} +  \frac{1}{2}\nabla_{q} \left(\dot{q}^{T} M \dot{q}\right).
\end{align}
If the system is fully actuated, one obtains expressions for the system's inverse dynamics by simply rearranging both \eqref{eq:fd_cranmer} and \eqref{eq:fd_lutter}, reading 
\begin{align}
    \Fu &= \left(\nabla_{\dot{q}}\nabla_{\dot{q}}^{\top} \mathcal{L}\right) \ddot{q} + \left(\nabla_{q}\nabla_{\dot{q}}^{\top} \mathcal{L}\right) \dot{q} - \nabla_{q} \mathcal{L} - \Fd, \label{eq:cranmer_inverse}\\
    &= M \ddot{q} + \nabla_{q} (\dot{q}^{\top}M) \dot{q} -  \frac{1}{2}\nabla_{q} \left(\dot{q}^{T} M \dot{q}\right) + \nabla_{q} V - \Fd. \label{eq:id_lutter}
\end{align}

\begin{figure}[t]
     \centering
     \begin{subfigure}[b]{0.45\textwidth}
         \centering
         \includegraphics[width=\textwidth]{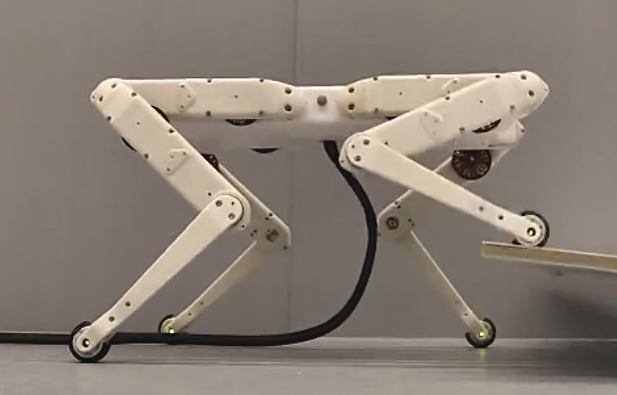}
     \end{subfigure}
     \hspace{0.5cm}
     \begin{subfigure}[b]{0.35\textwidth}
         \centering
         \includegraphics[width=\textwidth]{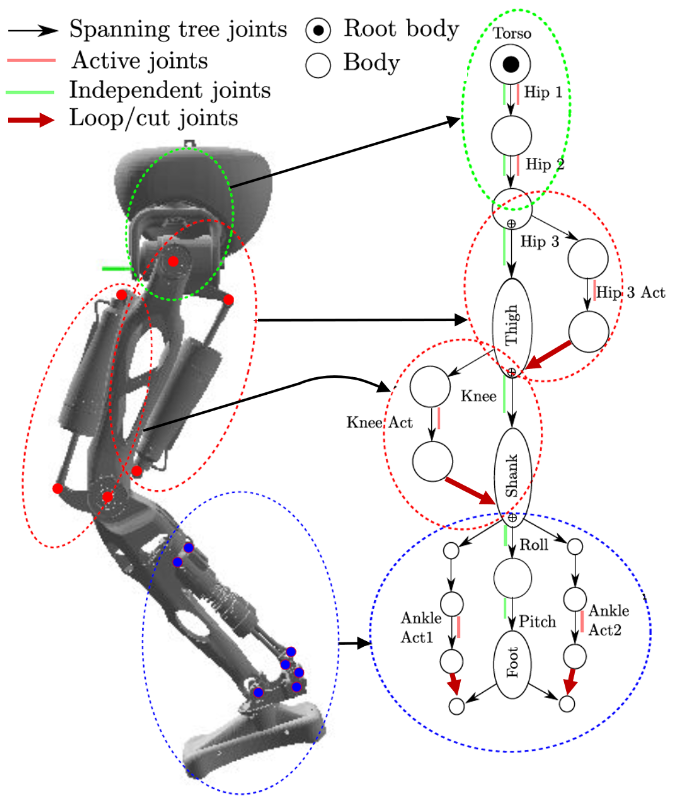}
     \end{subfigure}
     \caption{The feet of a quadruped (Left: Open Dynamic Robot, used with courtesy of \cite{grimminger2020open}) are subject to contact forces. A pneumatic-actuated leg (Right: RH5 leg, adapted with courtesy of \cite{kumar2019modular}) has several kinematic loops that can be modeled via implicit constraints which induce constraint forces at the cut joints.} \label{fig:quadruped}
\end{figure}

\subsection{Implicit constraint equations: Structural knowledge on the direction of motion} \label{sec:constraint_mechanics}
Section \ref{sec:mechanics_newtonian} outlined how constraint forces $F_{\text{E},i}$ can be excluded from the EOM by a suitable choice of generalized coordinates $q$ which yield explicit holonomic constraints. 
In this section, we emphasize the utility of implicit constraints as a complement to using explicit constraints and an additional source of structural knowledge. 

Implicit constraints allow to include constraint forces into the EOM. Implicit constraints can be particularly useful for non-holonomic systems, system's with kinematic loops, and systems that are subject to inequality constraints. For example, when the foot of a quadruped (Figure 2, left) presses onto a surface, the surface applies a reaction force that reduces the DOF of the system. The force arising from the contact can be modelled using an implicit holonomic constraint. In return, one obtains an analytical expression for the respective constraint force which can be straightforwardly removed from the EOMs if the constraint is inactive.
As another example, a hydraulic-actuated robot leg (Figure 2, right) introduces kinematic loops
that can be modeled via the addition of implicit constraints.
%
We base the following discussion on the EOM as in \eqref{eq:newton_euler}. However, we broaden the problem setting compared to Section \ref{sec:mechanics_newtonian} by making the following assumptions:
\begin{itemize}
    \item The system of rigid bodies is subject to constraint forces that now impose $n_{\text{E}}+n_{\text{I}}$ independent constraint equations with $n_{\text{I}}$ denoting the number of implicit constraints. In return, the system has $(n_q - n_{\text{I}})$ DOF \cite[p.\,43]{layton2012principles}.
    \item Of these constraint forces, $n_{\text{E}}$ are eliminated from the EOM through use of explicit holonomic constraints and a suitable choice of generalized coordinates $q\in \mathbb{R}^{n_q}$ as discussed in Section \ref{sec:mechanics_newtonian}.
\end{itemize}
These assumptions cover by no means all possible descriptions of the EOM of a rigid-body system. For example, constraint equations can be redundant which in return requires a more extensive treatment on the connection between forces and the vector spaces that constraint-related matrices span. A more general discussion is given in \cite{featherstone2014rigid} as well as \cite{koganti2016unified}. Nevertheless, the following discussion illustrates the interplay between many of the building blocks that are frequently encountered when deriving dynamics equations and which can yield structure to a regression model as detailed in Section \ref{sec:latent_constraint_modelling}.

 Implicit constraints can be expressed algebraically as $c(q,t)=0$ if they are holonomic, or more generally as $c(q,\dot{q},t)=0$ if they are non-holonomic. We assume that differentiation with respect to time yields implicit constraint equations on the system's position, velocity, and acceleration as detailed below in \eqref{eq:holonomic_implicit} and \eqref{eq:nonholonomic_implicit}.
\begin{alignat}{3}
&\text{position}  &&\text{velocity}  & & \text{acceleration} \nonumber\\
\text{holonomic:} \hspace{0.5cm}& c(q,t)=0 \hspace{0.5cm} \overset{\text{d}/\text{d}t}{\rightarrow} \hspace{0.5cm} && \tilde{A}(q,t)\dot{q} = \tilde{b}(q,t) \hspace{0.5cm} \overset{\text{d}/\text{d}t}{\rightarrow} \hspace{0.5cm} & & A(q,t)\ddot{q}=b(q,\dot{q},t) \label{eq:holonomic_implicit}\\
\text{nonholonomic:} \hspace{0.5cm} & \hspace{0.5cm}  && c(q,\dot{q},t)=0 \hspace{1cm} \overset{\text{d}/\text{d}t}{\rightarrow}  & & A(q,\dot{q},t)\ddot{q} = b(q,\dot{q},t) \label{eq:nonholonomic_implicit}
\end{alignat}
The terms such as $A(q,t)=\tilde{A}(q,t)=\frac{\partial c(q,t)}{\partial q}$ and $b(q,\dot{q},t) = \frac{\partial c(q,\dot{q},t)}{\partial t}$ denote partial derivatives with respect to $q$ or $t$, respectively. Comparing   \eqref{eq:holonomic_implicit} and \eqref{eq:nonholonomic_implicit}, one sees that  holonomic and non-holonomic constraints can be denoted jointly on the acceleration level as
    \begin{equation} \label{eq:constraining_equation}
        A\ddot{q}=b,
    \end{equation}
where we assume that only $n_{\text{I}}$ constraints are expressed in implicit form such that $A\in\mathbb{R}^{n_{\text{I}} \times n_q}$ and $b\in\mathbb{R}^{n_{\text{I}}}$.

\begin{remark}{Minimal state description of non-holonomic systems}
\newline
\emph{
To obtain a minimal state description, a non-holonomic system requires more position variables than velocity variables. For example, a unicyclist riding on a plane can reach any potential position on the plane. Yet, as an ideally rolling wheel cannot slide sideways, the wheel's translational Cartesian velocities at the contact point with the plane can be described in terms of a single velocity variable pointing along a position-dependent axis. However, in this work, we limit the discussion on the description of the system's EOM using as many velocity variables $\dot{q}$ as position variables $q$. An introduction to the derivation of EOMs of non-holonomic systems with a minimal state representation is given in \cite[p.\,41]{featherstone2014rigid} and \cite[p.\,114]{schiehlen2014applied}.
}
\end{remark}

\paragraph{Reformulation of virtual displacements}
To be able to eliminate constraint forces using the D'Alembert-Lagrange principle for multibody systems \eqref{eq:principle_of_virtual_work}, one must define what constitutes a virtual displacement vector $\delta q$. Oftentimes, the non-holonomic implicit constraints are assumed Pfaffian taking the form $\tilde{A}(q,t)\dot{q}=\tilde{b}(q,t)$ with $\tilde{A}(q,t)$ and $\tilde{b}(q,t)$ not being partial derivatives of a position-level constraint. In this case, for holonomic and Pfaffian non-holonomic constrained systems the virtual displacement is often defined as the infinitesimal vector $\delta q$ that fulfills
$\tilde{A} \delta q = 0$ (cf. \cite[p.\,131]{udwadia2007analytical} and \citep[p.\,50]{layton2012principles}). However, this definition is not applicable for non-holonomic constraints of the form in \eqref{eq:nonholonomic_implicit}. In this case, many works resort to virtual velocity vectors which denote the infinitesimal change in the velocity that agrees with the constraints while not varying position and time \citep[p.\,55]{schiehlen2014applied}. In return, one can turn to Jourdain's principle of virtual power to discuss the effect of non-holonomic constraint forces on the EOM. However, as our previous discussion of explicit constraints was centered around the concept of virtual work, we instead define virtual displacements as proposed by \citet{udwadia1997equations} as an arbitrary infinitesimal vector $\delta q \in \mathbb{R}^{n_q}$ fulfilling the equation
\begin{equation} \label{eq:virtual_displacements_A}
    A \delta q = 0.
\end{equation}
Unlike Section \ref{sec:mechanics_newtonian}, in which $\delta q$ denoted any infinitesimal vector inside $\mathbb{R}^{(6N_{\text{b}}-n_\text{E})}$, 
\eqref{eq:virtual_displacements_A} implies that in the presence of the additional implicit constraint forces, the virtual displacement denotes any infinitesimal vector $\delta q\in \Nspace(A)$. 
Note that the above definition of $\delta q$ is similar to the definition of virtual velocities as found in other textbooks (cf. \citep[p.\,202]{woernlemehrkorpersysteme}).

\begin{remark}{Outline of the derivation of \eqref{eq:virtual_displacements_A}}
\newline
\emph{
As shown by \citet{udwadia1997equations}, \eqref{eq:virtual_displacements_A} follows from the Taylor expansion about time $t$ of a displacement $q(t+dt)$ as
\begin{equation} \label{eq:taylor_q}
    q(t+dt) = q(t) + (dt)\dot{q}(t) + \frac{(dt)^2}{2}\ddot{q}(t) + \mathcal{O}(dt^3),
\end{equation}
with $dt$ being an infinitesimal quantity.
With \eqref{eq:taylor_q} one defines the virtual displacement at time (t+dt) as the difference between a possible displacement $q^b(t+dt)=\text{vec}\{q,\dot{q},\ddot{q}^b\}$ and the actual displacement $q^a(t+dt)=\text{vec}\{q,\dot{q},\ddot{q}^a\}$ such that
\begin{equation}
    \delta q = q^b(t+dt) - q^a(t+dt) = \frac{(dt)^2}{2}[\ddot{q}^b(t)-\ddot{q}^a(t)+\mathcal{O}(dt)].
\end{equation}
As every possible motion must fulfill \eqref{eq:constraining_equation}, we can insert $q^a(t+dt)$ and $q^b(t+dt)$ into \eqref{eq:constraining_equation} and take the difference between the expressions to obtain
\begin{equation}
    A(q(t), \dot{q}(t),t) [\ddot q^b(t) - \ddot q^a(t)]=0,
\end{equation}
which with $\eqref{eq:taylor_q}$ can be shown for $dt \rightarrow 0$ to yield  \eqref{eq:virtual_displacements_A}.
}
\end{remark}

\paragraph{D'Alembert's principle and implicit constraint forces}
The implicit constraints are the consequence of constraint forces and torques $F_{\text{I},i}=\text{vec}\{f_{\text{I},i}, \tau_{\text{I},i}\}$ acting onto each body. In turn, the D'Alembert-Lagrange principle for the multibody system \eqref{eq:principle_of_virtual_work} with the explicit transformation to generalized coordinates \eqref{eq:virtual_transfo} as well as $\sum_{i=1}^{N_\text{b}}J_i^\transp F_{\text{E},i} = 0$ due to the specific choice of $q$, reads
\begin{equation} \label{eq:Daelmbert_Q}
\sum_{i=1}^{N_\text{b}} \dv_i^\transp (F_{\text{E},i} + F_{\text{I},i}) = \delta q^\transp \sum_{i=1}^{N_\text{b}} J_i^\transp (F_{\text{E},i} + F_{\text{I},i}) 
= \delta q^\transp \FI = 0,
\end{equation}
with 
the generalized implicit constraint forces $Q_I = \sum_{i=1}^{N_\text{b}} J_i^\transp F_{\text{I},i}$.
%
As \eqref{eq:virtual_displacements_A} requires $\delta q \in \Nspace(A)$, \eqref{eq:Daelmbert_Q} yields that
    \begin{equation} \label{eq:dalembert_ideal}
        \FI \in \Rspace(A^\transp).
    \end{equation}
In turn, \eqref{eq:dalembert_ideal} motivates to parametrize the ideal constraint force in terms of a Lagrange multiplier vector $\lambda(q, \dot{q}, t) \in \mathbb{R}^{n_\text{I}}$ as
\begin{equation} \label{eq:lagrange_multipliers}
     \FI = A^{\top} \lambda.
\end{equation}
By inserting \eqref{eq:lagrange_multipliers} to the EOM in \eqref{eq:newton_euler} or \eqref{eq:euler_lagrange} one obtains an index-3 system of differential algebraic equations (cf. \citep[p.\,105]{schiehlen2014applied}) as
\begin{equation}\label{eq:index3}
    \begin{bmatrix} M & -A^\transp \\ A & \boldsymbol{0} \end{bmatrix} \begin{bmatrix} \ddot{q} \\ \lambda \end{bmatrix} = \begin{bmatrix} \F \\ b\end{bmatrix}.
\end{equation}

\begin{figure}[t]
     \centering
       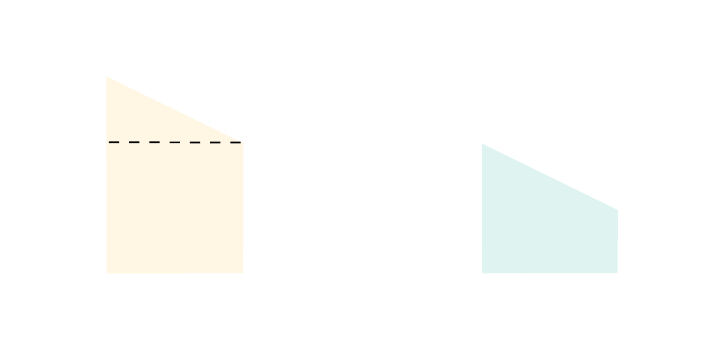
 \caption{Under the assumption that the explicit and implicit constraints are independent, 
 the constraint forces and virtual displacements must lie in the null and range spaces of $J,J^\transp,A,$ and $A^\transp$ as illustrated above. In the above diagram being inspired by \citet{beard2002linear}, vertical lines denote vector spaces. One can move a vector between two of these spaces by multiplication from the left with the constraint-related matrix transformation indicated on the respective arrow.} 
 \label{fig:constraint_space_diagram}
\end{figure}

\paragraph{Non-ideal constraint forces}
Oftentimes, constraint forces are not ideal and do virtual work. In such a case, \citet{udwadia2002foundations} proposed to divide the constraint force into an ideal part $\FI$ and a non-ideal part $\Fdi$ doing virtual work. The forces $\Fdi$ are dissipative and therefore can be seen as being part of $\Fd$ as long as the respective constraints are active. Oftentimes, $\FI$ causes $\Fdi$. For example, the friction force $\Fdi$ between a robot's foot and a surface is caused by the normal force $\FI$ that the foot applies onto the surface. 
  As emphasized by \citet{udwadia2000nonideal}, the part of any arbitrary force which is doing virtual work, $\F^{\prime}$, writing $\delta q^\transp \F^{\prime}\neq 0$, must have the same direction as $\delta q$ and hence 
      \begin{equation} \label{eq:dalembert_nonideal}
       \F^{\prime} \in \Nspace(A).
    \end{equation}

\paragraph{ODE form of the EOM for implicitly constrained systems}
Many different dynamic formulations for $\FI$ in terms of implicit constraints and the unconstrained dynamics equations have been proposed in literature. For example, \citet{aghili2005unified} details several dynamics equations of implicitly holonomic constrained systems. 
Alternatively, for the case of independent implicit constraints and a positive-definite mass matrix, applying the matrix inversion lemma on \eqref{eq:index3} yields the description of the EOM in ODE form
    \begin{align} \label{eq:UKE11}
        \ddot{q} &= M^{-1} \Big( \F + \underbrace{A^T(AM^{-1}A^T)^{-1}(b - AM^{-1}\F}_{\text{\normalsize $\FI$}}) \Big).
    \end{align}
\citet{udwadia2006explicit} showed that \eqref{eq:UKE11} is a special case of the so called Udwadia-Kalaba equation, which yields EOM for systems with positive \emph{semi}-definite inertia matrix and dependent implicit constraint equations. An introduction to the usage of the Udwadia-Kalaba equation for robot control is given in \citep{peters2008unifying}.
Note that $\FI=A^T(AM^{-1}A^T)^{-1}(b - AM^{-1}\F)$ yields an explicit form for the Lagrange multiplier parametrization in terms of the constraining equation \eqref{eq:constraining_equation}.
Rearranging \eqref{eq:UKE11} yields
\begin{align} \label{eq:uke_rewritten}
    \ddot{q} &= M^{-1} \left( P\F + \F_b \right),
\end{align}
with the weighted constraint projection $P(q,\dot{q},t) = I_n - A^T(AM^{-1}A^T)^{-1}AM^{-1}$ and the constraining force $\F_b = A^T(AM^{-1}A^T)^{-1}b$. 
As further discussed by \citet{udwadia1992new}, Gauss \cite{gauss1829neues} observed that the acceleration caused by an ideal constraint force, $\ddot{q}_I=M^{-1}\FI$, minimizes the quadratic functional
\begin{equation} \label{eq:gauss_principle}
        G(q,\dot{q}, t)=\ddot{q}_I^{\top} M \ddot{q}_I.
    \end{equation}
The above equation, being referred to as \emph{Gauss' principle of least constraint}, uniquely defines the length of the vector $\ddot{q}_{\text{I}}$ in \eqref{eq:UKE11}. In turn, the matrix $P$ is a weighted projection from the $n_q$-dimensional force space to $\Nspace(A)$ such that Gauss' principle is fulfilled, where $\Nspace(A)$ is a $(n_q-n_I)$-dimensional manifold inside $\mathbb{R}^{n_q}$.

\paragraph*{Vector spaces and constrained dynamics}
The assumption that the explicit and implicit constraint equations are independent requires $J^\transp$ and $A$ to have full row-rank at every non-zero state-vector $\{q, \dot{q}, t\}$ that respects the constraints. Under these assumptions, the insights on the direction of $F_{\text{E}}$ in \eqref{eq:space_of_Fe}, $\dv$ in \eqref{eq:space_of_dv}, $\delta q$ in \eqref{eq:virtual_displacements_A}, and $\FI$ in \eqref{eq:dalembert_ideal} can be summarized in a single diagram as depicted in Figure \ref{fig:constraint_space_diagram}\hspace{-0.18cm}. Figure \ref{fig:constraint_space_diagram}\hspace{-0.18cm} emphasizes that at an admissible point in the system's state-space $\{q, \dot{q}, t\}$, the virtual displacement vectors as well as constraint forces are bound to lie in spaces spanned by the constraint-related matrices $J$ and $A$. In addition, the inertia matrix $M$ defines the mapping from the acceleration space to the force space. Gauss' principle as in \eqref{eq:gauss_principle} emphasizes that $M$ also determines an analytical expression for the implicit constraint force $\FI$ as in \eqref{eq:UKE11}.


\section{Analytical structured learning} \label{sec:structured_learning}
In the following section, we propose a unified view on analytical structured modeling.  For this, we leverage that the dynamics descriptions presented in Section \ref{sec:mechanics} consist of sums of latent vector-valued functions (\eg forces) that are multiplied with matrix-valued functions (\eg the inertia matrix). We then show that this perspective enables us to decompose and discuss the error functions inherent in an analytical model. 
As data-driven models are a substantial part of analytical structured modeling, we then proceed by giving a brief introduction to common data-driven dynamics models and analytical models in parametric network form. Finally, we discuss selected literature on ARM in Section \ref{sec:analytical_output} and ALM in Section \ref{sec:analytical_latent}.

\subsection{A unified view on analytical model errors} \label{sec:unified}
To understand the pros and cons of different analytical structured models, we require a thorough understanding of the cause of the analytical model errors. As shown in Section \ref{sec:mechanics}, the forward dynamics can be expressed either in terms of force vectors \eqref{eq:newton_euler}, additional potential functions \eqref{eq:euler_lagrange}, or with additional implicit constraints \eqref{eq:uke_rewritten}. All of these formulations of the EOM consist of a sum of generalized forces $(\F+\FI)$ that are multiplied by $M^{-1}(q,\ta)$. In comparison, the inverse dynamics formulations in \eqref{eq:id_lutter} or \eqref{eq:cranmer_inverse} form sums of latent functions that are transformed by $M(q,\ta)$ or simply identity matrices. To unify the discussion, we therefore assume that rigid-body dynamics equations can be written as a sum of latent vector-valued functions $\hat{f}_i(x,\ta)$ that are transformed by latent matrix-valued functions $\C(x;\ta)$, such that the analytical model becomes
\begin{equation} \label{eq:linear_analytical_model}
   \hat{f}_{\text{A}}(x;\ta) = \sum_{i}\C\hat{f}_i.
\end{equation}
As analytical models of a mechanical system dynamics often erroneous, one can substitute \eqref{eq:linear_analytical_model} into \eqref{eq:analytical_approximation_error} to obtain an expression of the system's dynamics in terms of the analytical model and multiple error functions, writing
\begin{align} 
   f(x) &= \hat{f}_{\text{A}} + \epsilon_{\text{A}} = \hat{f}_{\text{A}} + \epsilon_{\text{E}} + \er, \label{eq:error_formulation1}\\
   &= \er + \sum_{i}\left(\C + \epsilon_{\mathcal{C},i}\right)\left(\hat{f}_i+\epsilon_{\hat{f},i}\right), \label{eq:error_formulation12}
\end{align}
with 
the vector-valued error functions $\er(x)$, $\epsilon_{\text{E}}(x)$, and $\epsilon_{\hat{f},i}(x)$, as well as the matrix-valued error function $\epsilon_{\mathcal{C},i}(x)$. 

%
\begin{figure}[t]
     \centering
             \includegraphics[trim=100 175 100 150,clip,width=0.8\textwidth]{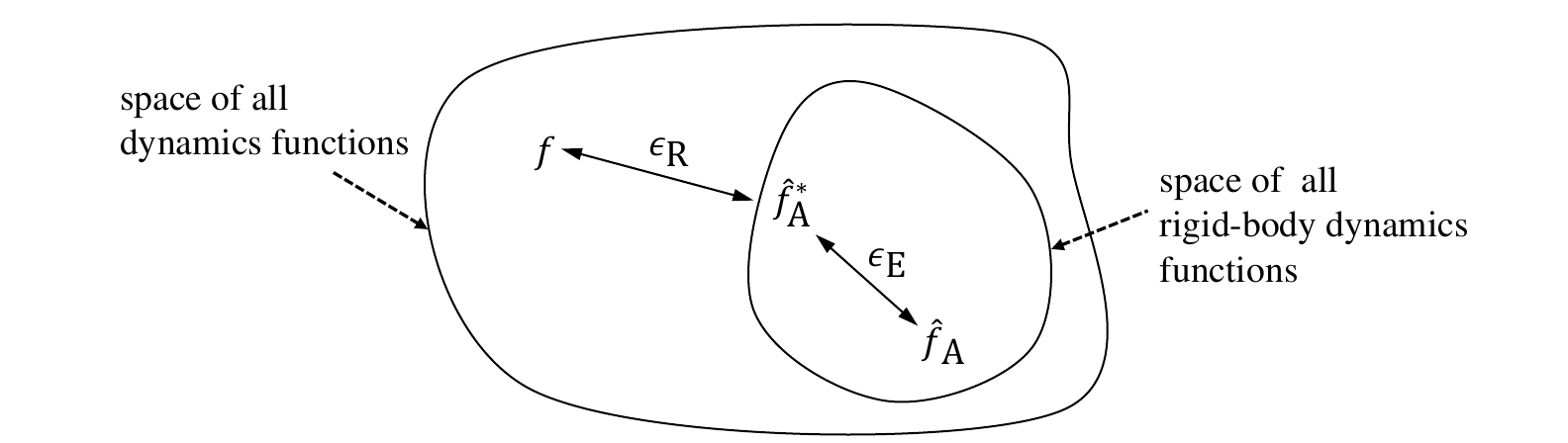}
         \caption{Illustration of the errors of a rigid-body dynamics model.}
         \label{fig:error_functions}
\end{figure} 

\begin{figure}[t]
     \centering
             \resizebox{135mm}{!}{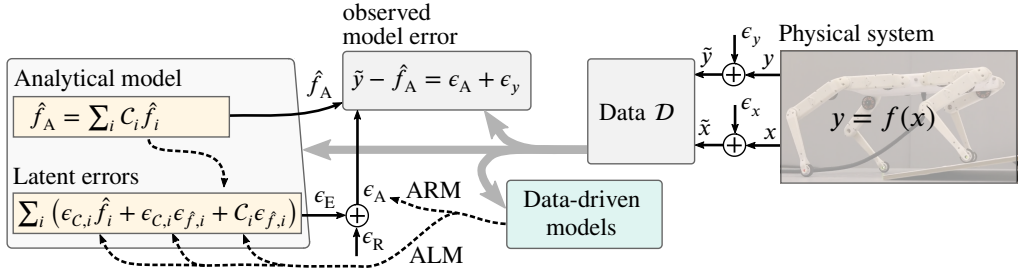}
         \caption{Schematic of analytical structured modeling. Data-driven models can approximate the output error of the analytical model (ARM) and/or reduce latent errors inside the analytical model (ALM). Robot image with courtesy of \cite{grimminger2020open}.}
         \label{fig:overview}
\end{figure}

\paragraph{Ideal models and latent error functions}
 To shed further light on the error functions in \eqref{eq:error_formulation1} and \eqref{eq:error_formulation12}, we define the \emph{ideal rigid-body dynamics model} 
\begin{equation} \label{eq:optimallinear_analytical_model}
   \hat{f}_{\text{A}}^* (x;\ta^*) = \hat{f}_{\text{A}} + \epsilon_{\text{E}} = \sum_{i}\C^{*}\hat{f}_i^*,
\end{equation}
which corresponds to the analytical model in which the asterisk symbol $(^*)$ denotes adequate choices of $\C$,$\hat{f}_i$, and $\ta$ such that $\|\er(x)\|$ is minimized. With this definition, $\er$ denotes all errors that cannot be captured using rigid-body dynamics modelling. For example, $\er$ can be caused by elasticities in the system's bodies or disturbances caused by attached cables. In contrast, $\epsilon_{\hat{f},i}$ and  $\epsilon_{\mathcal{C},i}$ cause errors between the rigid-body dynamics model $\hat{f}_{\text{A}}$ and the ideal rigid-body dynamics model $\hat{f}_{\text{A}}^*$. Figure \ref{fig:error_functions}\hspace{-0.18cm}, being inspired by \cite{von2011statistical}, illustrates the errors between the system's dynamics $f$ and an analytical approximation $\hat{f}_{\text{A}}$. 
In turn, the analytical model error reads
\begin{equation} \label{eq:error_formulation2}
    \epsilon_{\text{A}} = \er + \epsilon_{\text{E}} =\er + \sum_{i}\left( \epsilon_{\mathcal{C},i}\hat{f}_i + 
    \epsilon_{\mathcal{C},i} \epsilon_{\hat{f},i} +
    \C \epsilon_{\hat{f},i} \right).
\end{equation}
The foremost goal behind the combination of data-driven models with an analytical model is to reduce $\epsilon_{\text{A}}$. Therefore, as depicted in Figure \ref{fig:overview}\hspace{-0.18cm}, analytical structured models can be distinguished by how $\epsilon_{\text{A}}$ is reduced using data-driven modeling, namely into:
\begin{itemize}
    \item ARM, in which data-driven models approximates $\epsilon_{\text{A}}$ directly,
    \item ALM of latent analytical functions, in which data-driven models approximate $\hat{f}_i$ and/or $\C$ in \eqref{eq:error_formulation1} if the respective analytical model is inaccurate, or alternatively,
     \item ALM of latent residual functions, in which data-driven models approximate  $\epsilon_{\mathcal{C},i}$ and/or $\epsilon_{\hat{f},i}$ in \eqref{eq:error_formulation1} hence also using the analytical functions $\hat{f}_i$ and $\C$.
\end{itemize}

\paragraph{Using the unified view to compare direct and inverse dynamics}
To illustrate the implications underlying \eqref{eq:error_formulation1}, consider the forward dynamics as in \eqref{eq:UKE11}. With \eqref{eq:error_formulation1} we then obtain
\begin{equation} \label{eq:errors_in_direct_euler}
       \ddot{q} = \er + \left(M^{-1} + \epsilon_{M^{-1}}\right) \left(\F+\FI +\epsilon_{\F}\right),
\end{equation}
where the error functions in the entries of $M^{-1}$ are denoted by $\epsilon_{M^{-1}}(q)$, and error functions in the entries of $\F+\FI \,\widehat{=}\, \sum_{i}\hat{f}_i$ are denoted by $\epsilon_{\F}(q,\dot{q},t)$.
%
In comparison, the system's inverse dynamics read
\begin{equation} \label{eq:errors_in_inverse_euler}
    \Fu = \er + \left(M +  \epsilon_{M}\right)\ddot{q} - \left(\FC + \FG + \Fd + \FI + \epsilon_{\text{CGd}}\right),
\end{equation}
with the force errors $\epsilon_{\text{CGd}}(q,\dot{q},t)$. 
 
A model deviates from the dynamics either trough model errors as in \eqref{eq:errors_in_direct_euler} and \eqref{eq:errors_in_inverse_euler} or, through observation noise.
The \emph{model errors} occurring in \eqref{eq:errors_in_inverse_euler} can be denoted jointly as $\er+\epsilon_{M} \ddot{q} + \epsilon_{\text{CGd}}$. As these errors directly occur in the output of the inverse dynamics they can straightforwardly approximated using ARM. In comparison, forward dynamics formulations as in \eqref{eq:errors_in_direct_euler}, nonlinearly transform the error $\epsilon_{\F}$ by $\left(M^{-1} + \epsilon_{M^{-1}}\right)$.
As illustrated in Example \ref{ex:latent_modelling}, this can render ARM significantly more challenging for forward dynamics compared to inverse dynamics.
%
\begin{example} \label{ex:latent_modelling}
\textbf{Structured modeling of pendulum dynamics} \\
\emph{The forward dynamics of an undamped-uncontrolled pendulum in terms of its angle $q$ reads
\begin{equation}
    \ddot{q}(q) =M^{-1} \Fg,
\end{equation}
 with the inverse of the inertia matrix $M^{-1} = 1/mL^2$ and the gravitational torque $\Fg=-\sin(q)Lmg$ consisting of the gravitational acceleration $g$, the pendulum's mass $m$, and the length of the pendulum's rod $L$. For the sake of this illustration, assume that the estimate for the gravitational acceleration $g$ is erroneous such that an a-priori available model reads $\hat{\F}_g = -\sin(q)Lm(g + \epsilon_g)$ where $\epsilon_g$ denotes a constant function. Even though $\epsilon_g$ is constant, if its propagated through the dynamics function the output prediction error becomes a nonlinear function
\begin{equation}
\epsilon_{\ddot{q}}(q) = \frac{-\sin(q)}{L} \epsilon_g.
\end{equation}
Therefore, if the model designer is confident that the inverse inertia matrix $M(q,m,L)^{-1}$ and the kinematic dependency $\sin(q)L$ are suiting parametrizations of the real system's physics, one can place a data-driven model on $\epsilon_{g}$ instead of $\epsilon_{\ddot{q}}$. In turn, a less complex data-driven model can be used. 
Alternatively, one can use the additional prior knowledge that $g>0$, such that $\hat{\F}_g=-\sin(q)Lm\hat{g}^2$ where $\hat{g}(x;\tm)$ denotes a suitable data-driven model.
}
\end{example}
Another significant quantity to consider is the \emph{acceleration noise}. Estimates of $\ddot{q}$ are usually obtained via numerical time-differentiation of velocity or position measurements. In return, the measurement noise is amplified in the acceleration estimate. As it is often easier for data-driven models to approximate the comparably large noise in the model's \emph{output} compared to the model's input, forward dynamics models can be preferred to inverse dynamics models. This is also a reason that might explain why the majority of data-driven models approximate transition dynamics \eqref{eq:transition_dynamics} as state measurements are usually readily available. However, as forward dynamics formulations are usually used to compute trajectory predictions via numerical integration methods, these trajectory predictions contain an additional integration error. 

Another important aspect forms the measurement of $\Fu$. Recall that the actuation dynamics $\Fu'(x, \F_{\text{u,desired}})$ denote the difference between $\F_{\text{u,desired}}$ and $\Fu$. The identification of $\Fu'$ is therefore critical if we want to use a dynamics model for control or simulation. However, the estimation of $\Fu$ via current measurements in electric motors is inaccurate. In return, if we learn an inverse dynamics model, the model's output error $\epsilon_y$ will also contain the difference between $\Fu$ and its estimation. Alternatively, we can use the end-effector force to estimate $\Fu$. However, such estimates depend on mechanical parameters.
In comparison, directly measuring $\Fu$ through force sensors in the actuators yields accurate measurements. Albeit, sensors, such as joint torque sensors, are costly. 
Compared to learning inverse dynamics with ARM, learning forward dynamics with ALM has the advantage that we do not need to measure $\Fu$ but instead approximate it directly with a data driven-model.

\paragraph{Combining ARM with ALM: An uncharted territory}
While we separate analytical structured modeling in between the lines of ARM and ALM, these two modeling approaches can be combined if the dynamics are affected by phenomenons that are not describable by rigid-body dynamics.

As shown in Section \eqref{sec:strcutured_forces}, some of the recent works that apply ALM on forward dynamics equations assume that $M^{-1}$ in \eqref{eq:errors_in_direct_euler} accurately depicts the causal map between the forces and acceleration of a system. Accordingly, the error $\epsilon_{M^{-1}}$ is only caused by a wrong estimate of $\ta$.  
Placing a data-driven model on the error-prone parts of $\F$ (ALM) as well as on the output residual of the structured model (ARM), potentially reduces $\epsilon_Q$ and $\er$. If the reduction in the former error terms causes $\ta$ to be closer to $\ta^{*}$, $\epsilon_{M^{-1}}$ also reduces. Consequently, 
the prediction performance significantly improves. However, this is currently solely a hypothesis based on the empirical validations made in the literature discussed in Section \ref{sec:analytical_latent}. To which extend modeling of forward dynamics benefits from a combination of ALM and ARM requires further future analysis.

\subsection{Data-driven dynamics models and analytical parametric networks}
In this section, we briefly discuss solely data-driven dynamics models as these models are important for structured modeling. Further, we discuss works that learn the parameters of analytical models using gradient-based optimization. The insights about analytical models are useful for improving future generations of analytical structured models.

\subsubsection{Data-driven dynamics models}
Common data-driven models used for the identification of dynamics are (Bayesian) linear regression, neural networks (NNs), and Gaussian processes (GPs). Most works that learn dynamics function using solely data-driven models use transition dynamics or inverse transition dynamics. \citet{nguyen2011model} surveys these approaches for robot control. 
In the following, we briefly discuss NNs and GPs as examples of data-driven models being used for analytical structured learning.

NNs achieved breakthroughs in big-data domains such as computer vision \citep{krizhevsky2012imagenet}, the game of go \citep{silver2016mastering}, and control of computer games \cite{berner2019dota,vinyals2019alphastar}. NNs have also been deployed for the inference of physical system dynamic's \citep{funahashi1993approximation, kuschewski1993application, jansen1994learning, morton2018deep}. Notably, NNs have learned successfully transition dynamics and control-policies on contact rich domains \citep{chua2018deep, nagabandi2020deep}. Remarkably, \citet{nagabandi2020deep} demonstrated that NNs enable the inference of specific control policies for handling objects inside a robotic hand. However, in \citep{nagabandi2020deep} the training of a control policy for a specific task-object combination requires several hours of data. 

In comparison, GPs are non-parametric and probabilistic models that define a normal distribution over functions. GPs provide a measure of uncertainty of the estimation result in form of their posterior variance. In addition, GPs convert Bayesian inference into numerically efficient linear algebraic equations. An introduction to multivariate GP regression is given in \cite{alvarez2011kernels}. In what follows, we denote a multivariate GP as $\hat{f}\sim \mathcal{GP}\left(\mathbf{m}(x),\mathbf{k}(x,x')\right)$ with the vector-valued mean function $\mathbf{m}(x)$ and matrix-valued kernel $\mathbf{k}(x,x')$.
\citet{deisenroth2011pilco} learned the system's transition dynamics with a GP. This GP dynamics model was then used for the nonlinear control of real mechanical systems with to NNs comparably small amounts of data. One key take away of \citep{deisenroth2011pilco} has been the usage of a probabilistic model which allows propagating uncertainty in the observed state-actions through the system dynamics. In turn, this approach improved significantly the robustness of a control policy trained on the GP dynamics model. Other applications of dynamics modeling with GPs for control includes
\citep{nguyen2010using,kocijan2005dynamic,frigola2013bayesian, mattos2016latent, doerr2017optimizing, eleftheriadis2017identification, doerr2018probabilistic}. The main disadvantage of GPs is their computational complexity, which typically scales cubically with the number of data points. Even though their computational complexity can be reduced using sparse GPs \cite{quinonero2005unifying}, the computational effort required to work with such models is considerably larger than using NNs.

\begin{table}[]
\centering
\caption{Summary of works on analytical structured models that are detailed in this survey. We use the abbreviation FD for forward dynamics and ID for inverse dynamics. The entries in the rightmost corner link to code repositories that have been submitted alongside the publications.} \label{tab:works_libraries}
\begin{tabular}{@{}l|lllllll@{}}
\toprule
& Publication & Year & Dynamics & \begin{tabular}[c]{@{}l@{}}Data-driven\\ model type\end{tabular} & \begin{tabular}[c]{@{}l@{}}Data-driven\\ approximation\end{tabular} & \begin{tabular}[c]{@{}l@{}}Real system\\ experiments\end{tabular}& Optimization library \\ 
\midrule
\multirow{1}{*}{APN} & Ledezma et al.\, \cite{ledezma2017first} & 2017 & ID \cite{an1985estimation} & $\times$ & $\times$ & \checkmark & \emph{fmincon} (Matlab)\\
& Ledezma et al.\,\cite{ledezma2018fop} & 2018 & ID \cite{an1985estimation} & $\times$ & $\times$ & \checkmark & \emph{fmincon} (Matlab)\\
& \citet{sutanto2020encoding} & 2020 & FD \cite{luh1980line} & $\times$ & $\times$ & \checkmark &\href{https://github.com/facebookresearch/differentiable-robot-model}{PyTorch (Python)} \\
\midrule
\multirow{1}{*}{ARM}
& Nguyen-Tong et al.\,\cite{nguyen2010using} & 2010 & ID & GP & $\epsilon_{\text{A}}$ & \checkmark & $\times$ \\
& \citet{de2011online} & 2011 & ID & LWPR & $\epsilon_{\text{A}}$ & $\times$ & $\times$ \\
& \citet{um2014independent} & 2014 & ID & GP & $\epsilon_{\text{A}}$ & $\times$ & $\times$  \\
& \citet{grandia2018contact} & 2018 & ID & GP/LWPR & $\tilde{\epsilon}_u$ & \checkmark & GPy (Python)\\
\midrule
\multirow{1}{*}{ALM} 
& \citet{cheng2015learning} & 2016 & ID & GP & $\mathcal{L}$ & \checkmark & $\times$ \\
& Geist et al. \cite{geist2020gp2} & 2020 & FD & GP & $\FI$ &  $\times$ & \href{https://github.com/AndReGeist/gp_squared}{scikit-learn (Python)}\\
& \citet{hwangbo2019learning} & 2019 & FD \cite{hwangbo2018per} & NN & $\Fu$ & \checkmark & $\times$ \\
& \citet{greydanus2019hamiltonian} & 2019 & FD & NN & $\mathcal{H}$ & \checkmark & \href{https://github.com/greydanus/hamiltonian-nn}{PyTorch (Python)} \\
& \citet{lutter2018deep} & 2019 & FD/ID & NN & $M$, $\FG$ & \checkmark & \href{https://github.com/milutter/deep_lagrangian_networks}{PyTorch (Python)} \\
& \citet{lutter2019deep} & 2019 & FD/ID & NN & $M$, $V$ & \checkmark & \href{https://github.com/milutter/deep_lagrangian_networks}{PyTorch (Python)} \\
& \citet{gupta2020structured} & 2020 & FD/ID & NN & $M$, $V$, $B$, $\Fd$ & $\times$ & \href{https://sites.google.com/stanford.edu/smm/}{PyTorch / Flux (Julia)}\\
& \citet{lutter2020differentiable} & 2020 & FD \cite{kim2012lie} & NN & $\Fu$ &
\checkmark & $\times$ \\
 & \citet{toth2019hamiltonian} & 2020 & FD & NN & $\mathcal{H}$ & $\times$ & $\times$ \\
& \citet{cranmer2020lagrangian} & 2020 & FD & NN & $\mathcal{L}$ & $\times$ & \href{https://github.com/MilesCranmer/lagrangian_nns}{JAX (Python)} \\
\bottomrule
\end{tabular}
\end{table}

\subsubsection{Analytical parametric networks}\label{sec:APN}
Recent works propose to formulate analytical models as analytical parametric networks (APN) that are trained with gradient-based optimization methods. While technically these kinds of models are analytical models, the techniques used are closely related to techniques from data-driven modeling which in return blurs the line between analytical and solely data-driven modeling. 

\citet{ledezma2017first} reformulated the inverse dynamics of a robot arm such that both the kinematic and dynamic parameters can be estimated via gradient-based optimization. In this approach, the inverse dynamics of the robot arm are separated into a kinematic and dynamic network which allows estimating the kinematic parameters before estimating the dynamic parameters. This approach has been extended by \citet{ledezma2018fop} towards the identification of inverse dynamics of humanoid robots.

Similar to \cite{ledezma2017first}, \citet{sutanto2020encoding} described a recursive formulation of the Newton-Euler inverse dynamics as a differentiable computational graph. The dynamics parameters are estimated via automatic differentiation. In this work, the authors placed special emphasis on the incorporation of additional structural knowledge contained in the mass parameters as discussed in \cite{traversaro2016identification}. \citet{traversaro2016identification} show that not every positive-definite matrix constitutes a physically plausible inertia matrix as the rotational inertia matrix must also fulfill \emph{triangular inequalities} with respect to the principal moments of inertia. These insights led \citet{sutanto2020encoding} to a parametrization of the rotational inertia matrix of each body respecting triangular inequalities.
It should be noted that the experiments detailed in \cite{sutanto2020encoding} show similar training results for the models with parametrization of the rotational inertia matrix either solely in terms of positive definiteness or by taking the triangular inequality property into account.

\subsection{Analytical output residual modeling} \label{sec:analytical_output}
In this section, we discuss selected literature on ARM. In practice, in ARM a data-driven model simply approximates the analytical residual function $\epsilon_{\text{A}}+\epsilon_y$. 

\paragraph{From linear-regression of analytical models to semi-parametric ARM models}
Historically, ARM originated from the need for accurate dynamic models on robot arms. Here, \citet[p.\,280]{siciliano2010robotics} referrers to \cite{atkeson1986estimation} as the standard approach used for the identification of the dynamics parameters of robot arms. \citet{atkeson1986estimation} leverage the \emph{linearity} of the (rigid) robot arm's inverse dynamics with respect to the dynamical parameters \cite[p.\,259]{siciliano2010robotics}, writing 
\begin{equation} \label{eq:linear_dynamics_arm}
     \Fu = \Phi(q,\dot{q},\ddot{q})  \theta_{\text{A}}.
\end{equation}
The least squares estimate of $\theta_{\text{A}}$ is obtained as
\begin{equation} \label{eq:linear_arm}
    \hat{\theta}_{\text{A}} = (\Phi^{\top}\Phi)^{-1} \Phi^{\top} \tilde{\F}_u,
\end{equation}
with measurements of $\Fu$ being denoted as $\tilde{\F}_u$ and $(\Phi^{\top}\Phi)^{-1} \Phi^{\top}$ being the left pseudo-inverse matrix of $\Phi$. Some of these parameters denote linear combinations of dynamical parameters  \cite[p.\,280]{siciliano2010robotics}. However, the least-squares approach requires a good prior model of the kinematic parameters and dynamic parameters (masses and inertias) acting on the systems. Therefore, \citet{nguyen2010using} proposed to combine \eqref{eq:linear_arm} with GP regression resulting in a semi-parametric or fully-parametric structured model. The semi-parametric modeling approach simply approximates the analytical model's residual via a zero-mean GP, writing $\hat{\epsilon}_{\text{A}} \sim \mathcal{GP}\left(0,\mathbf{k}(x,x')\right)$ such that 
\begin{equation} \label{eq:duy_model1}
         \hat{\F}_u \sim \Phi(x)  \hat{\theta}_{\text{A}} + \mathcal{GP}\left(0,\mathbf{k}(x,x')\right) = \mathcal{GP}\left( \Phi(x)  \hat{\theta}_{\text{A}},\mathbf{k}(x,x')\right),
\end{equation}
with $\mathbf{k}(x,x')$ denoting a diagonal matrix-valued kernel function. A Gaussian distributed estimate of the system's dynamic parameters $\hat{\theta}_{\text{A}}$ can then be inferred via the posterior distribution of \eqref{eq:duy_model1}, writing $\hat{\F}_u | \mathcal{D}$  \cite[p.\,27-29]{williams2006gaussian}. 

The second model proposed in \cite{nguyen2010using} uses the kernel trick to obtain an analytical kernel of the inverse dynamics', writing
\begin{equation}
    \mathbf{k}_{\text{A}}(x,x') = \Phi^{\top}W\Phi + \Sigma_y,
\end{equation}
with $\Sigma_y$ denoting a diagonal matrix of observation noise variances and $W$ being a diagonal matrix denoting the prior variance on $\theta_p$. An algorithm that is defined solely in terms of inner products in input space is lifted by the kernel trick into feature space \cite[p.\,12]{williams2006gaussian}.
The analytical kernel GP can then be combined with $\hat{\epsilon}_{\text{A}} \sim \mathcal{GP}\left(0,k(x,x')\right)$ to yield
\begin{equation} \label{eq:duy_model2}
         \hat{\F}_u = \mathcal{GP}\left(0,k_{\text{A}}(x,x') + k(x,x')\right).
\end{equation}
\citet{nguyen2010using} showed that \eqref{eq:duy_model1} and \eqref{eq:duy_model2} achieved comparable prediction accuracy on a real robot arm while \eqref{eq:duy_model1} was slightly faster in computing predictions.
Similar to \cite{nguyen2010using}, \citet{de2011online} combined prior analytical knowledge of a robot arm's inverse dynamics with locally weighted parametric regression (LWPR) such that its receptive field is a first-order approximation of the analytical model. \citet{de2011online} assumed the analytical parameters as fixed. Other semi-parametric models for learning a robot arms inverse dynamics are found in \cite{camoriano2016incremental}.

\paragraph{Combining a recursive Newton-Euler formulation with GP regression}
 \citet{um2014independent} extended the semi-parametric model proposed in \cite{nguyen2010using}. In this work, the authors emphasize that for general kinematic trees -- such as robot arms -- one can obtain the system's inverse dynamics in form of a recursive formulation of the Newton-Euler equations \cite{siciliano2010robotics}. 
 The model assumes an accurate prior analytical model whose parameters are assumed known.  The proposed modeling procedure splits into a forward and backward pass computation scheme. First, via assumed knowledge of the kinematic and inertial parameters, the joint velocities as well $\hat{Q}_{\text{C}}$ and $\hat{Q}_{\text{g}}$ at each joint are computed in a forward pass through the kinematic tree. Secondly, in a backward pass, the joint torques residuals of each joint are computed using the respective joint velocity as well as the generalized force acting on the parent joints. The generalized force acting on the parent joints is itself a prediction by another GP. To pass knowledge recursively from joint to joint as well as using only a subset of relevant states as GP inputs are both important propositions. 

\paragraph{Using kinematics to derive a contact-invariant formulation of errors}
\citet{grandia2018contact} learned the residual of a quadruped robot's inverse dynamics in a formulation that stays invariant under \emph{changes in the contact configuration}. Here, if the point-feet of the quadruped are pressed onto the surface, constraints are activated which are expressed via \eqref{eq:constraining_equation}. As first step, the authors eliminate $\FI$ from the quadrupeds dynamics using D'Alemberts principle \eqref{eq:dalembert_ideal} such that
\begin{equation} \label{eq:grandia_eq1}
    (I-A^+A)\hat{\epsilon}_{\text{A}} = (I-A^+A) (\Fd +\Fu + \FC - M \ddot{q}).
\end{equation}
Secondly, while the constraint is active, every part of a force pointing in $\Rspace(A^+)$ causes a reaction force $\FI$ which in return can cause a jump in \emph{every} component of $\ddot{q}$. As learning the jump in $\ddot{q}$ is difficult, \citet{grandia2018contact} use a coordinate transformation $J_u$ to express the force error via
\begin{equation}
\epsilon_{\text{A}} = A^{\top} \tilde{\epsilon}_c + J_u \tilde{\epsilon}_u
\end{equation}
with $A^{\top} \tilde{\epsilon}_c \in\Rspace(A^+)$ and $J_u \tilde{\epsilon}_u\in\Nspace(A)$. A careful choice of $J_u$ ensures that the activation of constraints only changes the error $\tilde{\epsilon}_c$.
Therefore, with \eqref{eq:grandia_eq1} one obtains $(I-A^+A)\hat{\epsilon}_{\text{A}} = (I-A^+A)J_u \tilde{\epsilon}_u$ and in return a \emph{constraint invariant formulation} of the force error as
\begin{equation}
    \tilde{\epsilon}_u = \left((I-A^+A)J_u\right)^+ (I-A^+A) (\Fd +\Fu + \FC - M \ddot{q}).
\end{equation}
The authors approximated $\tilde{\epsilon}_u$ using LWPR as well as GP regression.
\newpage
\subsection{Analytical latent modeling} \label{sec:analytical_latent}
In analytical models, it is often known which of its latent functions are prone to modeling errors.  
ALM seeks to reduce the error of an analytical model by placing data-driven models on the unknown latent functions of an analytical model. Further, ALM allows to incorporate prior knowledge on the mathematical properties of an analytical latent function into the design of its data-driven approximation. 
In the following, we detail different works on ALM. These works differ in which part of the analytical model is approximated with a data-driven model, namely: \emph{(i)} The entries of $M$, $M^{-1}$, or the Lagrangian function, \emph{(ii)} the entries of $\Fu$, or \emph{(iii)} the entries of $\F$ which are transformed using constraint knowledge. 

\subsubsection{Latent modeling using energy conservation} \label{sec:latent_energy_conservation}
The generalized inertia matrix $M$ is of great significance for rigid-body dynamics modeling. The fictitious force $\FC$ can be derived in terms of $M$, see \eqref{eq:fictitious_force}. Moreover, if $\Fg$ denotes the gravitational force then this force is also a function of the mass parameters. Furthermore, in forward dynamics, $M^{-1}$ is multiplied with $\F+\FI$ while in inverse dynamics the inertial force $M\ddot{q}$ has a significant impact on the final estimation results. In the works presented in the previous section, it is a common assumption that the inertia matrix and even its parameters are known a-priori. However, this can lead to large errors in $\epsilon_M$ / $\epsilon_{M^{-1}}$ and the part of $\epsilon_Q$ / $\epsilon_{CGd}$ that is caused by $\FC$. Therefore, recent works propose to approximate $\epsilon_M$ / $\epsilon_{M^{-1}}$ via a data-driven model by either directly modeling the entries of $M$ or by parametrization of the inertia matrix in terms of a Lagrangian. The works on Langrangian and Hamiltonian NN were inspired by the seminal work of \citet{chen2018neural} on neural differential equations.

\paragraph{Learning the Lagrangian}
Often analytical parametrizations of $M^{-1}$ and $\F_C$ are either not available or the effort required obtaining these is considered too large. Instead, one can express the system's dynamics in terms of the Lagrangian function $\mathcal{L}$. If the Lagrangian is not explicitly time-dependent one obtains an expression for the forward-dynamics in \eqref{eq:fd_cranmer} and for the inverse-dynamics in \eqref{eq:cranmer_inverse}. One of the first works that modeled the Lagrangian function via a GP is \citep{cheng2015learning}.  \citet{cheng2015learning} placed a GP prior on $\mathcal{L}$ writing $\hat{\mathcal{L}}\sim\mathcal{GP}(0,k(x,x'))$ and then transformed the GP prior by the operators in \eqref{eq:cranmer_inverse} to obtain a structured model for the inverse dynamics equation of a conservative system. However, it is currently not clear how to insert such an $\hat{\mathcal{L}}$ into the forward dynamics equation \eqref{eq:fd_cranmer} as efficient multi-output GP regression requires that $\hat{\mathcal{L}}$ is solely linearly transformed. In comparison, \citet{cranmer2020lagrangian} models $\mathcal{L}$ via a NN. In return, the authors obtain structured models both  for \eqref{eq:fd_cranmer} as well as \eqref{eq:cranmer_inverse}.

\paragraph{Learning the Hamiltonian}
Unlike the Newtonian and Lagrangian formulations of classical mechanics, Hamiltonian mechanics is rarely used for describing the motion of rigid-body systems. Yet, Hamiltonian dynamics is of utmost importance in other branches of mechanics such as quantum mechanics, celestial mechanics, and thermodynamics (cf. \cite{greydanus2019hamiltonian}). Hamiltonian mechanics is a reformulation of classical mechanics using the Legendre transform into $2n$ first-order ODEs in terms of position coordinates $q\in\mathbb{R}^n$ and a canonical impulse $p\in\mathbb{R}^n$, writing
\begin{equation}
    \dot{q}_i = \frac{\partial \mathcal{H}}{\partial p_i}, \hspace{1cm}  \dot{p}_i = - \frac{\partial \mathcal{H}}{\partial q_i},
\end{equation}
with $\mathcal{H}(q,p,t)$, $\mathcal{H}: \in\mathbb{R}^{2n} \rightarrow \mathbb{R}$ denoting the Hamiltonian function. Similar to the Lagrangian NN, \citet{greydanus2019hamiltonian} parameterized the Hamiltonian in the above equation via a NN. This model was extended by \citet{toth2019hamiltonian} using a generative NN structure which enables the inference of Hamiltonian dynamics from high-dimensional observations such as images.

\paragraph{Learning the inertia matrix}
As shown in \eqref{eq:general_lagrangian}, the kinetic energy of a rigid-body system is described in terms of the generalized inertia matrix $M$. Thereby, the forward dynamics \eqref{eq:fd_lutter} as well as inverse dynamics \eqref{eq:id_lutter} can be denoted in terms of $M$ instead of $\mathcal{L}$. However, it is inexpedient to directly model the function entries of $M$ using a data-driven model. As also discussed by \cite{traversaro2016identification}, the inertia matrix $M$ as derived in Section \ref{sec:mechanics_newtonian} must be positive definite. Therefore, \citet{lutter2018deep} proposed a parametrization of $M$ in terms of a lower triangular matrix $L(q,\dot{q})$ such that $\hat{M}=\hat{L}(q,\dot{q})\hat{L}(q,\dot{q})^T$ ensures that $\hat{M}$ is symmetric. In addition, the diagonal of $\hat{L}(q,\dot{q})$ is enforced to be positive such that all eigenvalues of $\hat{M}$ are positive (cf. Section \ref{sec:APN}). To do so, the output layer of the NN that forms the diagonal entries of $\hat{L}(q,\dot{q})$ uses a non-negative activation function such as ReLU or Softplus onto which a small positive number is added to prevent numerical instabilities. 
\citet{lutter2018deep} modeled the potential forces $\FG$ as well as non-conservative forces $\F_D$ jointly via a NN. The authors named the resulting structured model a \emph{Deep Lagrangian Neural Network (DELAN)}. \citet{lutter2019deep} showed that DELAN can be used for the energy-based control of a Furuta pendulum in which they used the fact that the conservative force can be written in terms of a NN parametrization of the potential energy function $\hat{V}(q)$, writing $\Fg=-\nabla_q \hat{V}(q)$.

\citet{gupta2020structured} extended DELAN by leveraging that the control force is affine in the control signal $u$, writing $\F_u= B(q)u$.
\citet{gupta2020structured} added NN parametrizations of $B(q)$ and $V(q)$, such that the deep Lagrangian network becomes
\begin{equation} \label{eq:gupta_model}
    \ddot{q}=(\hat{L}\hat{L}^T)^{-1}\left( - \nabla_{q} (\dot{q}^{\top}\hat{L}\hat{L}^T) \dot{q} +  \frac{1}{2}\left(\nabla_{q} \left(\dot{q}^{T} \hat{L}\hat{L}^T \dot{q}\right)\right)^{T} - \nabla_{q} \hat{V}(q) + \hat{B}(q)u + \hat{\F}_d \right).
\end{equation}

\subsubsection{Latent modeling of joint torques} \label{sec:strcutured_forces}
For many analytical descriptions of the system's dynamics, one can assume that some of the analytical latent functions form better approximations of the real physics than others. For example, for a robot arm, one can argue that the inertia matrix $M$, the fictitious force $\FC$, and gravitational force $\FG$ form good parametrizations of the respective physics. The parameters $\ta$ of these analytical functions are most likely unknown but can be estimated alongside the parameters of a data-driven model. In this case, the system would still respect energy conservation with respect to the model's energy $\hat{E}(q,\dot{q};\ta) = \frac{1}{2}\dot{q}^T\hat{M}(q;\ta)\dot{q} + \hat{V}(q,\dot{q};\ta)$ similarly to the structured Lagrangian models. Under this assumption, the majority of the errors $\epsilon_Q$ / $\epsilon_{CGd}$ in \eqref{eq:errors_in_direct_euler} and \eqref{eq:errors_in_inverse_euler} stem from an inaccurate description of the joint torques $\Fu$. The following works learn the dynamics of robots with motor torques being applied inside the robot's joints.

\paragraph{Direct identification of joint torques and combination with a contact model}
\citet{hwangbo2019learning} identified the joint torques $\Fu$ induced by a quadruped's electric motors via a NN \emph{a-priori}.
The authors used a simple control scheme to let a quadruped trot and meanwhile measured $\Fu$ using joint-torque sensors as well as position errors and velocities. Then a NN was trained to predict $\Fu$ given a sequence of position errors and velocities. In this manner, the authors learned the complete mapping of $\Fu$ including complex interlaced control routines of the motors (PD torque control, PID current control, field-oriented control) as well as transmission and friction disturbances. 

Afterward, the trained NN joint torque model is combined with a rigid-body simulation \cite{hwangbo2018per}. The simulator uses a hard contact model that respects Coulomb friction cone constraints. Then, a NN based reinforcement learning algorithm was trained to control the quadruped in simulation. 
Notably, the kinematic and mass parameters of the analytical model were randomly initialized to increase the robustness of the control policy during training.

The fact that the trained control policy achieved impressive results on the real quadruped, indicates that the gap between modeled and real dynamics can be bridged via randomization of physical parameters (body length and mass) of a good analytical model in combination with a prior identification of $\Fu$. This work was further extended by \citet{lee2020learning} who trained an \href{https://www.youtube.com/watch?v=9j2a1oAHDL8}{unprecedented robust control policy} for a quadruped robot traversing challenging terrain.

\paragraph{Modeling of joint torques inside forward dynamics}
\citet{lutter2020differentiable} (being the authors of DELAN) combined Newton-Euler dynamics in Lie Algebra form \cite{kim2012lie} with a NN parametrization of $\Fu$. The authors compared several models for $\Fu$, with $f_{\text{NN}}$ denoting a NN model, namely
\begin{align}
    \text{Viscous:\:}& \hat{\F}_{\text{u}} \,\hat{=}\, \F_{\text{u,desired}} - \theta_v \dot{q}, \label{eq:friction_model_1}\\
    \text{Stribeck:\:}& \hat{\F}_{\text{u}} \,\hat{=}\, \F_{\text{u,desired}} - \operatorname{sign}(\dot{q})\left(f_{s}+f_{d} \exp \left(-\theta_s\dot{q}^{2}\right)\right)-\theta_v \dot{q}, \label{eq:friction_model_2}\\
    \text{NN Friction:\:}& \hat{\F}_{\text{u}} \,\hat{=}\, \F_{\text{u,desired}} - \operatorname{sign}(\dot{q})\left\|f_{\mathrm{NN}}\left(q, \dot{q}\right)\right\|_{1}, \label{eq:friction_model_3}\\
    \text{NN Residual:\:}& \hat{\F}_{\text{u}} \,\hat{=}\, \F_{\text{u,desired}} - f_{\text{NN}}(q,\dot{q}), \label{eq:friction_model_4}\\
    \text{FF-NN:\:}& \hat{\F}_{\text{u}} \,\hat{=}\, f_{\text{NN}}(\F_{\text{u,desired}}, q,\dot{q}). \label{eq:friction_model_5}
\end{align}
Equations \eqref{eq:friction_model_1},\eqref{eq:friction_model_2}, and \eqref{eq:friction_model_3} are guaranteed to be solely \emph{dissipative}, while the more classical NN parametrization in \eqref{eq:friction_model_4} and  \eqref{eq:friction_model_5} do not respect energy dissipativity. However, the first three models assume that the internal motor control routines do not cause an overshoot such that the real $\Fu$ is actually larger than $\F_{\text{u,desired}}$. 

The different structured forward dynamics models were trained on data from simulated and real pendulums. Additionally, these models where compared to NN black-box modeling as well as the linear regression model denoted in \eqref{eq:linear_dynamics_arm} and \eqref{eq:linear_arm} (cf. \cite{atkeson1986estimation}). The training results show that a random initialization of the link parameters compares similarly to having a good prior knowledge of the link parameters. This indicates that it is possible to learn analytical and NN parameters \emph{jointly} inside a structured model. Further, the joint torque models which enforce dissipativity of $\hat{\F}_{\text{u}}$ gave significantly better long-term predictions. The long-term predictions were computed by feeding the models' acceleration predictions to a Runge-Kutta-4 solver.

\subsubsection{Latent modeling using implicit constraint knowledge} \label{sec:latent_constraint_modelling}
For some mechanical systems, such as a robot arm whose end-effector touches a surface or a quadruped robot walking over terrain, it can be desirable to identify the force acting in an implicit constraint directly from data. As detailed in Section \ref{sec:constraint_mechanics}, the presence of implicit constraint forces $\FI$ potentially also causes a non-ideal force $\Fdi$. While $\Fdi$ can be modeled jointly with $\Fd$, the implicit constraints as in \eqref{eq:constraining_equation} allows to place additional prior knowledge onto the direction of the force vectors. 
\newline\newline
To the best of our knowledge, \citet{geist2020gp2} were the first that proposed the usage of constrained knowledge for ALM. Here, the authors assumed that an analytical parametrization of $M^{-1}$, and the terms of the constraining equation $A$, and $b$ is given. The parameters of these analytical functions are estimated alongside the data-driven model's parameters. 
In \cite{geist2020gp2}, a GP prior is placed onto $(M^{-1}\F) \sim \mathcal{GP}(0,k(x,x'))$ inside \eqref{eq:uke_rewritten}, that is, the acceleration that would be caused by $\F$ if the constraints were absent. However, it is more straightforward to place a prior on $\F$ rather than its acceleration $M^{-1}\F$. With this small adaptation, the model obtained in \cite{geist2020gp2} reads
\begin{equation}
    \hat{\ddot{q}} \sim \mathcal{GP}(M^{-1}\F_b+(M^{-1}P)m_{\F},\: (M^{-1}P) k_{\F}(M^{-1}P)^{\top}),
\end{equation}
with analytical mean function $m_{\F}(x)$ and  kernel function $k_{\F}(x,x')$. Here, one can include analytical prior knowledge of $\FG$ and $\FC$ via the analytical mean function, writing $m_{\F} = \FG + \FC + \Fu$. If such a prior mean function is chosen, $k_{\F}(x,x')$ models $\Fd$ as well as the residual of $m_{\F}$. While the combination of \eqref{eq:uke_rewritten} with a parametric model is straightforward, using a GP has several advantages. For example, one can denote the joint distribution between $\F$ and $\ddot{q}$, writing
\begin{equation} \label{eq:joint distribution2}
   \begin{bmatrix} \hat{\F} \\ \hat{\ddot{q}} \end{bmatrix} \sim \mathcal{GP} \begin{pmatrix} \begin{bmatrix}  m_{\F}\\ M^{-1}\F_b+(M^{-1}P)m_{\F} \end{bmatrix}, \begin{bmatrix} k_{\F} &  k_{\F} (M^{-1}P)^{\transp} \\
   (M^{-1}P) k_{\F} & (M^{-1}P)k_{\F}(M^{-1}P)^{\transp}\end{bmatrix} \end{pmatrix}.
\end{equation}
This allows inferring $\hat{\F}$ from observations of $\ddot{q}$. Secondly, in GP\textsuperscript{2} one can learn on one constraint configuration $\{A,b\}$ and then change to a different constraint configuration $\{A',b'\}$ without the need for retraining if both constraints induce the same dissipative force function $\Fdi$ inside the constraint.

Besides directly combining \eqref{eq:uke_rewritten} with a regression model, one could also use projections into the constraint related vector spaces as depicted in Figure \ref{fig:constraint_space_diagram}\hspace{-0.18cm} without including the inertia matrix. In particular, the projection of a vector in $\mathbb{R}^{n_q}$ into either $\Nspace(A)$ and $\Rspace(A^\transp)$, (cf. \cite[p.\,74]{beard2002linear}), are given by $P^{\Nspace(A)}$ and $P^{\Rspace(A^\transp)}$.

\section{Key techniques for analytical structured modeling} \label{sec:synthesis}
 In order to design an analytical structured model, the following steps are required:
 \begin{enumerate}
     \item Derive an rigid-body dynamics model $\hat{f}_{\text{A}}(x;\ta)$.
\item Obtain a structured model $\hat{f}(x,\theta)$, by the combination of data-driven models with the analytical model. 
\item Collect informative data and estimate the parameters of the structured model.
 \end{enumerate}

If the parameters of an analytical model are assumed to be known, one can simply obtain an ARM model by learning $\epsilon_{\text{A}}+\epsilon_y$ using a data-driven model. However, ARM becomes significantly more challenging if $\ta$ is to be learned jointly with $\tm$. This case is particularly interesting as reducing $\epsilon_{\text{A}}$ potentially improves the estimate of $\ta$. 

In this section, we detail key techniques for the design and training of analytical structured models, in which $\ta$ and $\tm$ are estimated jointly as $\theta = \{\ta, \tm\}$. The first key technique is the optimization algorithm itself. Note that, all works in Table \ref{tab:works_libraries}\hspace{-0.18cm} use gradient-based optimization. Therefore, in Section \ref{sec:gradient_chapter}, we discuss gradient-based optimization and illustrate why automatic differentiation is particularly useful to compute the gradient of structured models. By discussing the optimization method, we see which requirement the optimization poses onto the design of an analytical structured model. Subsequently, we detail in Section \ref{sec:model_design} important aspects that must be considered when combining analytical models with data-driven models.

\newpage
\subsection{Gradient-based optimization} \label{sec:gradient_chapter}
Analytical structured models are used if standard analytical or data-driven approaches yield unsatisfactory predictions. Usually, this occurs if the system is high-dimensional and subject to nonlinear physical phenomenons. Therefore, most works utilizing analytical structured models require the training of high-dimensional models on large datasets.
\footnote{A dynamics model of a quadruped robot usually has more than 14 output dimensions (\eg Six DOF for the floating base plus two DOF per leg). In this scenario, a large dataset can denote just several thousands of points.}
Subsequently, recent works on analytical structured learning predominantly use gradient-based optimization algorithms to compute the model's parameters.

Gradient-based optimization forms a cornerstone of solely data-driven modeling \cite{bottou2018optimization} and constitutes one of the most common nonlinear local optimization techniques \cite[p.\,90]{nelles2013nonlinear}. Table \ref{tab:works_libraries}\hspace{-0.18cm} summarizes the literature presented in this survey that utilizes gradient-based optimization. Gradient-based optimization techniques require the computation of the partial derivative $\nabla_{\theta}\ell \, \widehat{=} \, \frac{\partial \ell}{\partial \theta}$ of an objective function $\ell(\hat{f}(\tilde{x};\theta),\tilde{y}) \, \widehat{=} \, \ell(\theta)$, $\ell \in \mathbb{R}$.
Often it is useful to include higher-order derivatives of the objective function in the parameter update rule. As gradient-based optimization is a local optimization technique and $\ell(\theta)$ is usually non-linear, it is indispensable to restart the optimization several times with random initialization of $\theta$, keeping only the best optimization result. In addition, as models based on NN have a large number of parameters, the gradient update is computed only for a randomly selected batch of data. This optimization procedure is called stochastic gradient-descent in machine learning literature.  
 
A typically used objective function for NN models is the $L_2$ loss function with an $L_2$ regularization term on the network's weights. For GP models the objective function is commonly chosen to be the logarithmic likelihood function. 
Both of these objective functions denote forward functions mapping $\theta$ to a cost $\ell(\theta)$.
 
As detailed by \citet{baydin2017automatic}, gradients of a forward function can be obtained via: 
\begin{itemize}
\item \emph{Analytical derivation and implementation}, which is time-consuming and error-prone. 
\item \emph{Numerical differentiation}, which is inaccurate due to round-off and truncation errors and scales poorly with the size of $\theta$.
\item \emph{Symbolic differentiation} yielding functions for the analytical gradient. However, these expressions are also closed-form which hinders GPU acceleration and quickly become cryptic due to an "expression swell" \cite{corliss1988applications}.  
\item \emph{Automatic differentiation}, which computes accurate gradients, is considerably faster than the aforementioned methods, and thanks to steady improvements in the usability of optimization libraries is straightforward to use with ALM.
\end{itemize}

\paragraph{Structured learning with automatic differentiation}
At the core of automatic differentiation (AD), which is also called algorithmic differentiation or ''autodiff``, lies the insight that forward functions are compositions of elementary operations whose derivatives are known. In return, the derivative of a function can be constructed using the elementary operators' gradient expressions with the chain rule of differentiation. AD computes a numerical value of the derivative of a computational graph with branching, recursions, loops, and procedure calls \cite{baydin2017automatic}. 

\begin{figure}[b]
     \centering
             \includegraphics[trim=100 180 100 150,clip,width=0.8\textwidth]{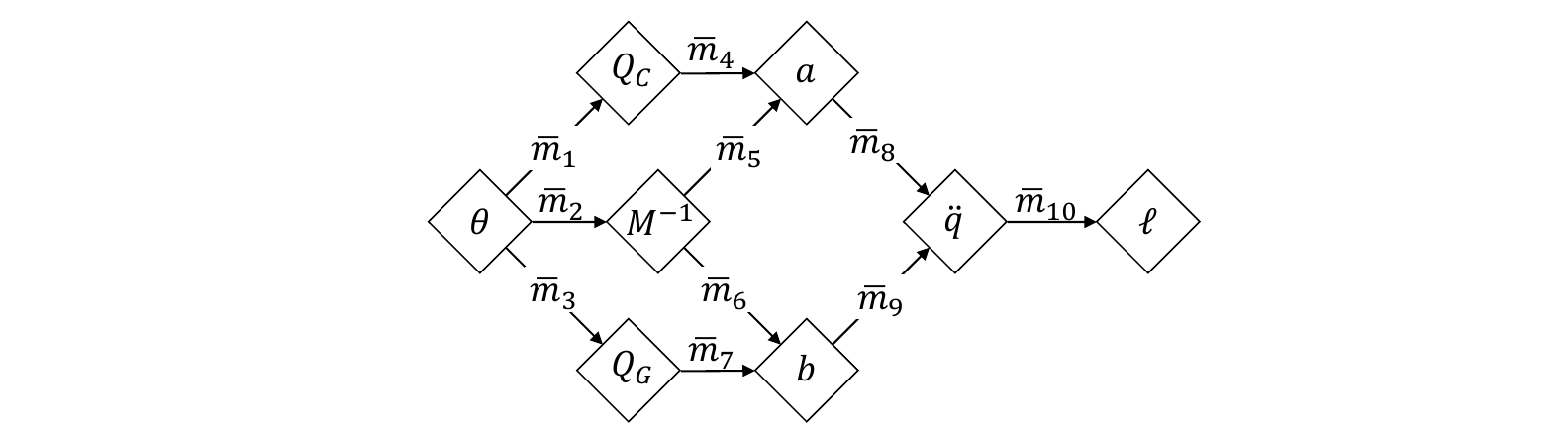}
         \caption{Graph illustrating backward mode automatic differentiation of the objective function of a forward dynamics model.  
         }
         \label{fig:backprop}
\end{figure} 

The two basic forms of AD are forward-mode AD and reverse-mode AD. In what follows, we denote the objective function as $\ell(\theta)=\ell(c(b(a(\theta))))$ where $a(\theta)$, $b(a)$, and $c(b)$ are functions of appropriate size. Then, in \emph{forward-mode AD}, the gradient of a function is obtained via forward accumulation, \eg writing $\nabla_{\theta}\ell(\theta) = \nabla_{c}\ell(\nabla_bc(\nabla_a b \nabla_{\theta}a))$ where $\nabla_{\theta} b = \nabla_a b \nabla_{\theta}a$ denotes a Jacobian matrix. 

Alternatively, \emph{reverse-mode AD} being also referred to as ''backprop`` in machine learning literature, computes gradients via reverse accumulation, \eg writing $\nabla_{\theta}\ell(\theta) = ((\nabla_c \ell \nabla_bc)\nabla_ab)\nabla_{\theta}a$. For functions $\ell:\mathbb{R}^{n_1} \rightarrow \mathbb{R}^{n_2}$ with $n_1 >> n_2$ reverse mode AD is preferred as it requires less operation counts for the computation of vector-Jacobian products \cite{baydin2017automatic}. However, reverse-mode AD can have increased memory requirements compared to forward-mode AD. Many automatic differentiation packages allow to combine forward and reverse accumulation.
In the following example, we illustrate reverse-mode AD for ALM.

\vspace{0.1cm}
\begin{example} \label{ex:backprop}
\textbf{Computing the gradient of a structured model with reverse-mode AD} \\
\emph{
Assume that the dynamics function of a mechanical system is modeled via a structured analytical model as
\begin{equation} \label{eq:example2}
    \hat{\ddot{q}}(x;\theta) = \underbrace{M^{-1}(x;\theta) \hat{\F}_{\text{C}}(x;\theta)}_{\text{\normalsize $a$}} + \underbrace{M^{-1} (x;\theta)\hat{\F}_{\text{G}}(x;\theta)}_{\text{\normalsize $b$}},
\end{equation}
in which $\hat{\F}_{\text{G}}(x;\theta)$ and $\hat{\F}_{\text{C}}(x;\theta)$ denote either analytical and/or data-driven models, and intermediate function expressions are abbreviated as $a=M^{-1}\hat{\F}_{\text{C}}$, $b=M^{-1}\hat{\F}_{\text{G}}$. Note that Figure \ref{fig:backprop}\hspace{-0.18cm} depicts the computational graph formed by \eqref{eq:example2}. The parameters $\theta$ are estimated jointly using an objective function of the form $\ell(\hat{f}(x;\theta),\tilde{y}) \, \hat{=} \, \ell(\theta)$. 
In reverse mode AD, the gradient of $\ell$ with respect to $\theta$ is decomposed as
\begin{equation}
   \nabla_{\theta} \ell(\theta) = 
    \underbrace{\nabla_{\hat{\F}_{\text{C}}} \ell \nabla_{\theta} \hat{\F}_{\text{C}}}_{\text{\normalsize $\bar{m}_1$}}
    + \underbrace{\nabla_{M^{-1}} \ell \nabla_{\theta} M^{-1}}_{\text{\normalsize $\bar{m}_2$}}
     + \underbrace{\nabla_{\hat{\F}_{\text{G}}} \ell \nabla_{\theta} \hat{\F}_{\text{G}}}_{\text{\normalsize $\bar{m}_3$}}, 
     \end{equation}
with
\begin{align}
    &\bar{m}_1 = \underbrace{\nabla_{a} \ell \nabla_{\hat{\F}_{\text{C}}} a}_{\text{\normalsize $\bar{m}_4$}} \nabla_{\theta} \hat{\F}_{\text{C}},
     \hspace{0.5cm}
     \bar{m}_2 = \big(\underbrace{\nabla_{a} \ell \nabla_{M^{-1}} a}_{\text{\normalsize $\bar{m}_5$}} +
     \underbrace{\nabla_{b} \ell \nabla_{M^{-1}} b}_{\text{\normalsize $\bar{m}_6$}}\big)
     \nabla_{\theta} M^{-1},
    \hspace{0.5cm}       
    \bar{m}_3 = \underbrace{\nabla_{b} \ell \nabla_{\hat{\F}_{\text{G}}} b}_{\text{\normalsize $\bar{m}_7$}} \nabla_{\theta} \hat{\F}_{\text{G}}, 
    \\
    &\bar{m}_4 = \underbrace{\nabla_{\hat{\ddot{q}}} \ell \nabla_{a} \hat{\ddot{q}}}_{\text{\normalsize $\bar{m}_8$}} \nabla_{\hat{\F}_{\text{C}}} a,
     \hspace{0.5cm}
    \bar{m}_5 = \underbrace{\nabla_{\hat{\ddot{q}}} \ell \nabla_{a} \hat{\ddot{q}}}_{\text{\normalsize $\bar{m}_8$}} \nabla_{M^{-1}} a,
     \hspace{0.5cm}
     \bar{m}_6 = \underbrace{\nabla_{\hat{\ddot{q}}} \ell \nabla_{b} \hat{\ddot{q}}}_{\text{\normalsize $\bar{m}_9$}} \nabla_{M^{-1}} b,
     \hspace{0.5cm}
    \bar{m}_7 = \underbrace{\nabla_{\hat{\ddot{q}}} \ell \nabla_{b} \hat{\ddot{q}}}_{\text{\normalsize $\bar{m}_9$}} \nabla_{\hat{\F}_{\text{G}}} b.
\end{align}
Therefore, the gradient of an analytical structured model can be computed with AD. Yet, this requires the computation of the numerical expressions for $\nabla_{\theta} M^{-1}$, $\nabla_{\theta} \hat{\F}_{\text{G}}$, and $\nabla_{\theta} \hat{\F}_{\text{C}}$. Ideally, an AD library will compute these terms for us by also decomposing the gradients into product-sums of basic gradient functions.
}
\end{example}

Example \ref{ex:backprop} illustrates that the computation of $\nabla_{\theta} \ell(\theta)$ requires the partial differentiation of analytical latent functions. Yet, to do this efficiently, several requirements must be fulfilled by an AD library, namely: 
\begin{itemize}
    \item Analytical structured models denote forward functions that consist of numerous basic mathematical operators. Ideally, the AD library should only require that the analytical structured model is expressed in terms of a computer library for basic linear algebraic operations such as Numpy \cite{harris2020array} or Torch \cite{NEURIPS2019_9015}. In turn, the model designer is only required to convert the derived analytical functions into the programming language of the respective AD library.
    \item As detailed in Section \ref{sec:latent_energy_conservation}, models for $\hat{\F}_{\text{C}}$, $\hat{\F}_{\text{G}}$, or $\hat{M}$ can be obtained by transformation of an analytical or data-driven function via the partial derivative operators $\nabla_q$ or $\nabla_{\dot{q}}$. Therefore, the AD library must be able to automatically compute the gradients (w.r.t. to $\theta$) of gradients (w.r.t. $q$ or $\dot{q}$) of latent functions.
\end{itemize}
Fortunately, recent developments in AD packages such as ''AutoGrad`` \cite{maclaurin2015autograd} and \href{https://pytorch.org/docs/stable/autograd.html}{''Torch autograd``} allow to compute $\nabla_{\theta}\ell(\theta)$ in which $\ell(\theta)$ is expressed in either native python (Numpy) code or python (torch) code. Importantly in these AD packages, $\nabla_{\theta}\ell(\theta)$ can itself include higher-order partial derivatives without braking the AD routines. These AD packages have been further improved in packages such as JAX \cite{jax2018github} and PyTorch \cite{NEURIPS2019_9015}, which additionally combine AD with compilers for GPU acceleration such as \href{https://www.tensorflow.org/xla}{XLA}. These packages tremendously simplify the synthesis of an analytical structured model, enables easy debugging of the respective computer code, as well as provide computationally efficient implementations with GPU acceleration.

\newpage
\subsection{Model construction of analytical latent models} \label{sec:model_design}
In the previous section, we detailed how an analytical structured model is trained using modern libraries for AD. These AD libraries only require that a forward function $\ell(\hat{f}(\tilde{x};\theta), \tilde{y})$ is expressed in terms of a specified programming language (\eg Python using solely Numpy expressions). In this section, we detail important aspects that must be considered when constructing an analytical structured model.
An analytical model $\hat{f}_{\text{A}}(x;\ta)$ can be derived using for example a rigid-body dynamics software package such as Pinocchio \cite{carpentier2019pinocchio}. 

If a data-driven model is used as an approximation for a function inside an analytical mode, this usually places additional requirements on the properties of the data-driven model. 
ALM requires placing data-driven models on unknown functions inside an analytical model. 
As discussed in Section \ref{sec:structured_learning}, these unknown quantities are either vector-valued functions $f_i$ (\eg forces) or matrix-valued functions $\C$ (\eg matrices).
Depending on the analytical mechanics formulation, these data-driven functions are usually transformed by operators which in return poses additional mathematical requirements onto the data-driven models. 
These transformations of data-driven models can be distinguished into:
\newline\newline
\noindent \emph{Matrix transformations}, writing
\begin{equation} \label{eq:linear_transformations}
    \hat{f}_i^{\mathcal{C}}(x) = \C(x) \hat{f}_i(x).
\end{equation}
\eg in forward dynamics a model can be placed on $\Fd$ which is transformed by $M^{-1}$, writing
$
    \hat{\ddot{q}}_{\text{d}} = M^{-1} \hat{F}_D.
$

\noindent \emph{Partial derivative transformations}, writing
\begin{equation} 
    \hat{f}_i^{\nabla_{x}}(x) = \nabla_{x} f_i(x)|_{x=z},
\end{equation}
which we denote by abuse of notation as
\begin{equation}
    \hat{f}_i^{\nabla_{x}}(x) = \nabla_x f_i(x),
\end{equation}
\eg modeling $\FC$ in terms of a model for the kinetic energy $\hat{T}$ (cf. \eqref{eq:fictitious_force}) reads
$
    \hat{\F}_{\text{C}}(q,\dot{q}) = - \left(\nabla_{q}\nabla_{\dot{q}}^{\top} \hat{T}\right) \dot{q} + \nabla_{q}\hat{T}.
$

\noindent \emph{Substitution of input variables by a nonlinear mapping} $x=u(z)$ such that
\begin{equation}
    \hat{f}_i^{u}(x) = f_i(z) |_{z=u(x)},
\end{equation}
\eg instead of using an angle coordinate $x \, \widehat{=} \,\phi$ for describing the pose of a pendulum, the pendulums pose can be expressed as $z=[\text{cos}(\phi), \text{sin}(\phi)]$. This coordinate transformation has the advantage that the entries of $z$ are bounded to $[-1,1]$. Another example is the descriptions of $z$ in terms of a NN with inputs $x$.
\newline\newline
\noindent Note that all of the above operators are linear operators. 
We illustrate the additional requirements that the above operators pose onto a data-driven model on the examples of NN and GPs. 

\paragraph{Linearly transformed neural networks}
\citet{hendriks2020linearly} details linear transformations of NNs such that an equality constrained is fulfilled. Additional details on ARMs with linear transformed NNs is found in \cite{lutter2018deep, lutter2019deep, gupta2020structured, cranmer2020lagrangian, greydanus2019hamiltonian, toth2019hamiltonian}  (cf. Section \ref{sec:analytical_latent}). If the NN outputs parameterize the entries of $M$, the positive definiteness of $M$ poses additional requirements on the last layer of the deep network. Moreover, transforming a NN with partial differential operators requires a careful selection of the activation functions. For example, the Lagrangian NN developed by \citet{cranmer2020lagrangian} contains the Hessian $\left(\nabla_{q}\nabla_{\dot{q}}^{\top} \hat{\mathcal{L}}\right)$ of the NN $\hat{\mathcal{L}}$. The second-order derivative of a RELU activation with respect to its inputs is zero. Therefore, the authors used and compared RELU\textsuperscript{2}, RELU\textsuperscript{3}, tanh, sigmoid, and softplus activation functions. For their Lagrangian NN the authors chose the softplus activation function. The usage of higher-order gradient-based optimization methods additionally affects the choice of the activation functions.

\paragraph{Linearly transformed Gaussian processes}
\citet{jidling2017linearly} and \citet{lange2018algorithmic} detail linear transformations of GPs such that an equality constrained is fulfilled.  Additional details on ARMs with linear transformed GPs is found in \cite{cheng2015learning, geist2020gp2}  (cf. Section \ref{sec:analytical_latent}). Gaussian processes are closed under linear operators, such that with $\hat{f}_i \sim \mathcal{GP}(\mathbf{m}(x), \mathbf{k}(x,x'))$, \eqref{eq:linear_transformations} yields
\begin{equation}
     \hat{f}_i^{\mathcal{C}} \sim \mathcal{GP}\left((\C(x)\mathbf{m}(x),\: \C(x)\mathbf{k}(x,x') \C^{\top}(x')\right),
\end{equation}
with the vector-valued mean function $\mathbf{m}(x)$ and matrix-valued (diagonal) kernel function $\mathbf{k}(x,x')$. The resulting GP $\hat{f}_i^{\mathcal{C}}$ is a multi-output GP \cite{MAL-036}. Most works that do ARM with GPs, model all output dimensions via independent GPs as the computation of the inverse covariance matrices of $n$ one-dimensional GPs scales with $\mathcal{O}\big(nN^3\big)$. In comparison, the computation of the covariance matrix of an $n$-dimensional multitask GP scales with $\mathcal{O}\big((nN)^3\big)$. However, a big advantage of multi-output GPs obtained from ALM modeling is that the transformation $\C(x;\theta)$ is often known a-priori from analytical mechanics. For example, in case of the Newton-Euler equation $\ddot{q}=M^{-1}\F$, we can place a GP prior on $\F$, writing $\hat{\F} \sim \mathcal{GP}(\mathbf{m}(x), \mathbf{k}(x,x'))$, such that
\begin{equation}
    \hat{\ddot{q}} \sim \mathcal{GP}\left((M^{-1}\mathbf{m},\: M^{-1}\,\mathbf{k}\,[M^{-1}]^{\top}\right).
\end{equation}
In turn, the prior knowledge of $M^{-1}$ specifies how the individual dimensions of $\hat{\F}$ are correlated with each other to yield $\hat{\ddot{q}}$.
As detailed in a \citet{jidling2017linearly} and \citet{lange2018algorithmic}, a GP can be transformed by a matrix of partial derivatives to yield another GP. For example, for $n=2$, the matrix of operators 
\begin{equation}
    \mathcal{C}_x = (\nabla_{q}\nabla_{\dot{q}}^{\top})=
    \begin{bmatrix}
    \frac{\partial^2}{\partial q_1 \partial \dot{q}_1} &  
    \frac{\partial^2}{\partial q_1 \partial \dot{q}_2} \\
    \frac{\partial^2}{\partial q_2 \partial \dot{q}_1} &  
    \frac{\partial^2}{\partial q_2 \partial \dot{q}_2}
    \end{bmatrix} 
\end{equation}
transforms a prior $\hat{\mathcal{L}} \sim \mathcal{GP}(\mathbf{m}(x), \mathbf{k}(x,x'))$ (cf. \cite{cheng2015learning}) as
\begin{equation}
    \mathcal{C}_x\hat{\mathcal{L}} \sim \mathcal{GP}(\mathcal{C}_x\mathbf{m}(x),\: \mathcal{C}_x\mathbf{k}(x,x')\mathcal{C}_x^{\top})).
\end{equation}
Therefore, if a GP is transformed by a matrix of partial derivatives, $\mathbf{m}(x)$ and $\mathbf{k}(x,x')$ must be sufficiently often differentiable with respect to the kernel's inputs. Further, as the sum of GPs are itself a GP (cf. \cite[p.\,13]{duvenaud2014automatic}), a model for $\F\,\widehat{=}\,\FG + \FC$ can be obtained as
\begin{equation}
    \hat{q} \sim \mathcal{GP}(\mathbf{m}_{\text{C}} + \mathbf{m}_{\text{G}},\: \mathbf{k}_{\text{C}} + \mathbf{k}_{\text{G}}\big),
\end{equation}
with $\hat{\F}_{\text{C}}\sim\mathcal{GP}(\mathbf{m}_{\text{C}}(x), \mathbf{k}_{\text{C}}(x,x'))$ and  $\hat{\F}_{\text{G}}\sim\mathcal{GP}(\mathbf{m}_{\text{G}}(x), \mathbf{k}_{\text{G}}(x,x'))$. 
\begin{figure}[t]
     \centering
       \resizebox{100mm}{!}{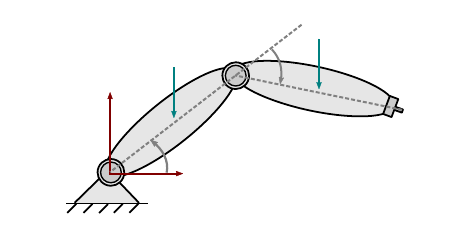}
 \caption{Sketch of a two-link robot arm who's end-effector touches a surface.}      \label{fig:robot_arm_diagram}
\end{figure}

\section{Case study: Structured modeling of a robot arm} \label{sec:example}
To illustrate the concepts presented in this survey, let us assume that we want to design an analytical structured model for a robot arm whose end-effector is in contact with a surface. We chose this example, as robot arms are widely used mechanical systems that yield complex dynamics functions. Moreover, a robot arm shares similar mechanical characteristics to a robotic leg which is an important component in robotics. While a 2D-system forms an abstraction of a real robot arm, similar modeling assumptions as in the 3D case are made while keeping the discussion concise. This discussion is based on the robot arm equations derived by \citet[p.\,265-269]{siciliano2010robotics}. Figure \ref{fig:robot_arm_diagram}\hspace{-0.18cm} depicts a schematic of the 2D robot arm. Here, Cartesian forces are denoted as $f$. The arm orientation is described via the generalized angle coordinates $q_1$ and $q_2$. 
As described in Section \ref{sec:mechanics}, the arm's rigid-body dynamics model takes the form
\begin{equation}
    \underbrace{M\ddot{q} = \FC + \FG}_\text{mass-related quantities} \: +\: \Fu + \underbrace{\Fd + \FI}_\text{constraint forces} .
\end{equation}
In this example, $\Fd\, \widehat{=} \, \Fdi$ denotes the friction force that is applied by the surface onto the end-effector while $\Fu$ denotes the joint torques.
While for all of these terms, analytical functions have been proposed, we need to ask ourselves if these models form good descriptions of the real system dynamics. As discussed in Section \ref{sec:structured_learning}, how we model the analytical functions heavily depends on the system and which quantities we can observe accurately.

\paragraph{Defining the analytical parameters}
It is assumed that the COG of the first body lies on an axis of length $a_1$ between the base joint and second joint. 
The COM of the second body lies on the axis of length $a_2$ between the second joint and the end-effector tip. The distance from the respective joints to a body's COM is denoted by $L_i$. The mass of each body is denoted by $m_i$ and the inertia around axis $e_3$ at the COM of the body by $I_{L_i}$. 
The robot arm is actuated by motors located in the joints. It is assumed that the COG of the motors is located at the origin of the respective generalized coordinate frames. The mass of the motor's rotors is $m_{m_i}$ and its inertia amounts to $I_{m_i}$. The gear reduction ratio of each motor is $k_{ri}$. 

\paragraph{Modeling the mass related quantities}
The inertia matrix of the robot arm follows from an analytical derivation as
\begin{equation}
    M(q) = \left[\begin{array}{ll}
b_{11}\left(q_{2}\right) & b_{12}\left(q_{2}\right) \\
b_{21}\left(q_{2}\right) & b_{22}
\end{array}\right] \text{ with } \: \begin{aligned}
b_{11}=& I_{L_{1}}+m_{L_{1}} L_{1}^{2}+k_{r 1}^{2} I_{m_{1}}+I_{L_{2}}\\
& + m_{L_{2}}\left(a_{1}^{2}+L_{2}^{2}+2 a_{1} L_{2} \text{cos}(q_2)\right) + I_{m_{2}}+m_{m_{2}} a_{1}^{2}, \\
b_{12}=& b_{21}=I_{L_{2}}+m_{L_{2}}\left(L_{2}^{2}+a_{1} L_{2} \text{cos}(q_2)\right)+k_{r 2} I_{m_{2}}, \\
b_{22}=& I_{L_{2}}+m_{L_{2}} L_{2}^{2}+k_{r 2}^{2} I_{m_{2}}.
\end{aligned}
\end{equation}
The fictitious force follows from \eqref{eq:fictitious_force} as
\begin{equation}
       \FC = - \nabla_{q}(\dot{q}^{\top}M) \dot{q} +  \frac{1}{2}\nabla_{q} \left(\dot{q}^{T} M \dot{q}\right) = - C(q,\dot{q}) \dot{q},
       \text{ with }\:
           C(q,\dot{q}) = \nabla_{q}(\dot{q}^{\top}M) =\left[\begin{array}{cc}
h \dot{q}_{2} & h\left(\dot{q}_{1}+\dot{q}_{2}\right) \\
-h \dot{q}_{1} & 0 
\end{array}\right],
\end{equation}
while the model for the gravitational force reads
\begin{equation}
   \FG = \left[
    \begin{array}{l}
\left(m_{\ell_{1}} \ell_{1}+m_{m_{2}} a_{1}+m_{\ell_{2}} a_{1}\right) g \text{cos}(q_1) +m_{\ell_{2}} \ell_{2} g \text{cos}(q_1+q_2) \\
m_{\ell_{2}} \ell_{2} g \text{cos}(q_1+q_2)
\end{array}\right],
\end{equation}
with $h=-m_{L_2}a_1 L_2 \sin(q_2)$.
To simplify this derivation, several assumptions were made which are also typically found in other analytical models, namely:
\begin{itemize}
    \item The COM of the bodies as well as of the motor's rotors lies on a predefined axis.
    \item Elasticities of bodies and compliance inside the joints are not modeled.
\end{itemize}
The latter point cannot be avoided using rigid-body modeling and hence can contribute to $\er$ (cf. Section \ref{sec:structured_learning}). To reduce $\er$ we can use ARM. However, learning these errors is difficult as we have only limited sensory information, \eg measurements of $q$, $\dot{q}$, and $\Fu$ ($\ddot{q}$ is often estimated by filtering measurements of $\dot{q}$ \cite[p.\,281]{siciliano2010robotics}). If we do not want to model $\er$ solely as a noise process, one can either take the history of the dynamics into account to learn on a streak of data points as done in \cite{hwangbo2019learning} or add additional sensors that capture information about the elasticities. 

Errors that stem from a wrong description of the bodies COM positions should ideally be avoided by improving the analytical model itself. It is tempting to directly resort to the data-driven methods presented in Section \eqref{sec:latent_energy_conservation} which model $M$, $\FG$, and $\FC$ via a NNs or GPs. However, for many mechanical systems, an accurate depiction of  $M$, $\FG$, and $\FC$ can be obtained from analytical modeling. The question is how much time and expertise are we willing to invest to obtain a sufficient analytical description of $M$, $\FG$, and $\FC$ and in return potentially improve the sample-efficiency of an analytical structured model.

\paragraph{Modeling the joint torques}
The identification of the non-conservative forces constitutes a big challenge in robotic systems. As discussed in Section \ref{sec:structured_learning}, how the joint torques $\Fu$ are modeled depends on how they are measured.

If the joint torques are directly measured through joint-torque sensors, one can directly learn the function $\Fu$. However, while theoretically, it should be possible to learn $\Fu$ from data of $\{q,\dot{q},\ddot{q}\}$, the acceleration $\ddot{q}$ is usually not directly observed and is therefore considerably noisy. In return, we suggest learning $\Fu$ using the techniques discussed in Section \ref{sec:strcutured_forces}. For example, $\Fu$ can be directly learned using a NN that has as input a sequence of position errors and velocities or $\F_{\text{u,desired}}$. 

Alternatively, if no accurate measurements of $\Fu$ are available one can design a structured inverse dynamics model as discussed in Section \ref{sec:analytical_output}. However, this model will learn the mapping from the current state-acceleration observation to the measured $\Fu$. In return, such a model is not necessarily suited for a simulation of the joint torques. A second approach is to directly learn the forward dynamics via ALM. By doing so, we need to only specify a model for $\Fu$.

\paragraph{Modeling the implicit constraint forces}
The identification of $\Fd$ and $\FI$ is also challenging as $\Fd$ is highly environmental dependent and $\FI$ is a contact force. We observed two different philosophies on how to model these forces in robotic systems. 

The first approach, which \citet{lee2020learning} used to achieve seminal results for the control a quadruped, is to directly measure and identify $\Fu$. Then we can insert the model for $\Fu$ in a rigid-body dynamics simulation where we can simulate different $\Fd$ and $\FI$. In return, one can design or learn a control strategy that yields robust performance over a large set of different $\Fd$. However, even in this scenario, it can desirable to identify a specific $\Fd$ as the robustness of a specific control policy to different dissipative forces can come at the cost of losing control performance.

The second approach, aims at modeling $\Fd$ inside the inverse or forward dynamics of the system. As shown in Section \ref{sec:constraint_mechanics} one can include constrained knowledge to eliminate $\FI$ from the dynamics equations. For example, assuming that the robot arm's end-effector is pressing on a flat surface such that the endeffector's Cartesian coordinate in $e_1$ direction denotes $r_{e,1} = 0$. Then, this surface is described by an implicit constraint equation as
\begin{align} \label{eq:duihdu}
     0 &= r_{e,1} = a_1 \text{cos}(q_1) + a_2 \text{cos}(q_1+q_2).
\end{align}
The the second time-derivative of \eqref{eq:duihdu} yields the constraining equation \eqref{eq:constraining_equation} as
\begin{align} \label{eq:duihdu2}
     \left[-a_1 \text{sin}(q_1) - a_2 \text{sin}(q_1+q_2), \: -a_2 \text{sin}(q_1+q_2)\right] \ddot{q} &= -a_1 \text{cos}(q_1) \dot{q}_1^2 - a_2 \text{cos}(q_1+q_2)(\dot{q}_1 + \dot{q}_2)^2.
\end{align}
This equation can be used to include constraint knowledge into an analytical structured model as detailed in Section \ref{sec:analytical_output} and Section \ref{sec:latent_constraint_modelling}.

\paragraph{Constructing the structured model and training}
Now we assume that all mechanical functions and error functions of the dynamics model are either modeled analytically or as a data-driven model.  As discussed in Section \ref{sec:gradient_chapter}, we suggest that these functions are then expressed in the programming language of a suitable automatic differentiation library to yield the analytical structured model. After the specification of the objective function, one can estimate the unidentified parameters of the structured model using collected data.

\section{Conclusion and future outlook}

Classical mechanics provides us with an understanding of the structural relationships underlying the motion of mechanical systems. One such insight is that motion is a consequence of the interaction of mass and force. Yet, to predict motion we need to describe the effects of mass and force mathematically. This turns out to be problematic in practice as the distribution of mass, as well as the applied forces, are itself the product of convoluted physical phenomena that are difficult, if not impossible, to describe a-priori. To address this and similar bottlenecks in a mathematical description of the world, we forged tools to learn functions from data. However, using data-driven learning methods comes at the price of requiring significant amounts of data as well as losing insight on how a prediction of such a model came about.

In this work, we discussed the different sources of structural knowledge inherent in analytical descriptions of rigid-body mechanics (Sec. 2). Here, we place special emphasis on the different type of forces, potential functions, and explicit and implicit constraint equations.

Further, we propose a unified view on structured modeling by emphasizing that output and latent modeling form key building blocks for analytical structured models (Sec. 3). In turn, we distinguish learning the residuals of an analytical model in the model's output, which is referred to as analytical (output) residual modeling (ARM), and analytical latent modeling (ALM), in which a data-driven model approximates latent residuals or functions inside the analytical model. To this effect, we consolidated the proposed view by illustrating that unknown analytical latent functions are often transformed by a known linear operator, which advocates the usage of ALM. 

In addition, our review of recent literature on analytical structured modelling (Sec. \ref{sec:structured_learning}), as well as related aspects of a model synthesis (Sec. \ref{sec:synthesis}), revealed a shift in the optimization techniques used in works on structured modeling. Namely, we observed a shift from analytical models that are optimized via linear regression towards structured latent models that are optimized via automatic differentiation. We believe that this shift 
is due to recent improvements in automatic differentiation libraries that significantly ease the synthesis of analytical structured models. 

Recently, an analytical structured model \cite{lee2020learning} enabled the synthesis of a quadruped robot's control policy with so far unseen and ineffable robustness to changes in the environment and carrying-load. 
We believe that this and the other works presented in this survey form merely the beginning of a broad application of analytical structured models in wide ranges of engineering, and we are excited to see what applications these promising class of models will enable in the near future.

\begin{center}\rule{0.4\textwidth}{.4pt}\end{center}

As analytical structured modeling is a relatively recent field, many open questions on the design of these models remain. Below we provide a list of what we perceive as potential directions that require further analysis in the field of analytical structured modeling, but that were beyond the scope of this survey.

\paragraph{Further types of rigid-body dynamics}
Many of the discussed works use the recursive Newton-Euler algorithm as thoroughly discussed in \cite{featherstone2014rigid}. However, some works also resorted to alternative dynamics descriptions such as the Newton-Euler equation in Lie algebra formulation \cite{kim2012lie} or the Udwadia-Kalaba equation \cite{udwadia2002foundations}. It seems that the recent works on analytical structured learning only touched the surface of the vast field of rigid-body simulations and it is worth further discussion which descriptions of motion are particularly suited for the combination with data-driven learning techniques. 

\paragraph{Partially observable systems}
There exist numerous works on data-driven modeling of partially observable systems \cite{lee2007makes, doerr2018probabilistic, curi2020structured}. Yet, works on analytical structured learning do typically not assume that the state is not directly observable and only partial state information is given via the observation equation $o_k = h(q^T, \dot{q}^T, u_k, \epsilon_o)$. Here $h(q^T, \dot{q}^T, u_k, \epsilon_o)$ most often denotes a simple algebraic equation and $\epsilon_o$ an additional noise process. The question remains if the existing insights from data-driven modeling can be extended to analytical structured modeling of partially observable systems.

\paragraph{Noise on model inputs}
The observations of the input variables $x$ are often noisy and biased. In dynamics modeling this is often not taken into account during training. The question arises if input noise can and should be addressed. Notably, some works propagate input noise through the trained dynamics model when doing predictions \cite{deisenroth2011pilco, parmas2018pipps}.

\paragraph{Active learning}
We assumed that informative system data is given. However, the collection of informative data from dynamical systems is itself a subject of ongoing research efforts \citep{schon2011system, simchowitz2018learning, schoukens2019nonlinear, buisson2020actively}. The question arises as to which extend structural knowledge can be used to make data-collection more efficient as well as how to use this data to train analytical structured models.

\paragraph{Interpretable data-driven models}
While data-driven models are often referred to as black-box models, there are efforts made to provide guaranties for data-driven model predictions \citep{umlauft2017learning, beckers2017stable} as well as carefully designing data-driven models to make them more interpretable \citep{ahmadi2020learning}. These works are critical if the application of a data-driven model requires prediction guarantees.

\paragraph{Combination of forward dynamics models with numerical integration}
As previously mentioned, a disadvantage of analytical structured learning of forward dynamics forms the noise on the acceleration estimates. Here, the forward dynamics are usually modeled to obtain a trajectory prediction using numerical integration techniques. An open question is to which extend analytical structured models can be combined with numerical integration schemes such that training of such a combined model is done solely on trajectory data. Recent works discuss how to approximate ODEs using GPs \cite{heinonen2018learning, yildiz2018learning, Wenketal20} and NNs \cite{chen2018neural, xuechen2020scalable} as well as solve initial-value problems using GPs \cite{schober2014probabilistic, schober2019probabilistic}.

\subsection*{Acknowledgments}
We thank Lucas Rath, Steve Heim, Alexander von Rohr, and Christian Fiedler for the valuable feedback and many interesting discussions. This work has been supported in part by the Max Planck Society and in part by the Cyber Valley initiative. The authors thank the International Max Planck Research School for Intelligent Systems (IMPRS-IS) for supporting A. René Geist.

\bibliography{literature} 


\end{document}